\documentclass{IEEEtran}
\title{Industry 4.0 and Prospects of Circular Economy: A Survey of Robotic Assembly and Disassembly}

\author{Morteza Daneshmand, Fatemeh Noroozi, Ciprian Corneanu, Fereshteh Mafakheri, Paolo Fiorini}
\usepackage[printonlyused]{acronym}
\usepackage[colorlinks=true]{hyperref}
\usepackage{graphicx}
\usepackage[T1]{fontenc}
\usepackage{amsmath}
\usepackage{subfigure}
\usepackage{multirow}
\usepackage{adjustbox}
\usepackage{cite}
\begin{document}
\maketitle

\begin{abstract}
Despite their contributions to the financial efficiency and environmental sustainability of industrial processes, robotic assembly and disassembly have been understudied in the existing literature. This is in contradiction to their importance in realizing the Fourth Industrial Revolution. More specifically, although most of the literature has extensively discussed how to optimally assemble or disassemble given products, the role of other factors has been overlooked. For example, the types of robots involved in implementing the sequence plans, which should ideally be taken into account throughout the whole chain consisting of design, assembly, disassembly and reassembly. Isolating the foregoing operations from the rest of the components of the relevant ecosystems may lead to erroneous inferences toward both the necessity and efficiency of the underlying procedures. In this paper we try to alleviate these shortcomings by comprehensively investigating the state-of-the-art in robotic assembly and disassembly. We consider and review various aspects of manufacturing and remanufacturing frameworks while particularly focusing on their desirability for supporting a circular economy.
\end{abstract}


\section{Introduction}

\begin{table*}[t]
\centering
\caption{A summary of the \ac{RAD}-related concepts reviewed throughout the paper, along with their linkages to the circular economy key performance indicators listed in Table~{20}, as well as the indices of the corresponding references and relevant sections.}
\begin{adjustbox}{max width=\textwidth}
\begin{tabular}{|c|c|c|c|c|c|c|c|c|c|}
\hline
Task & Subtasks & E1 & E2 & T1 & T2 & Eco1 & Eco2 & References & Section\\\hline
\multirow{9}{*}{Planning Robotic Assembly and Disassembly} & Classical Sequence Planning &  &  &  &  &  &  & \cite{warren2019computer} & \ref{sec:planning:classical_sp}\\\cline{2-10}
& Automatizing Sequence Planning &  &  & $\bullet$ &  &  &  & \cite{linnerud2019cad,nagasawa2019disassembly,said2019mathematic,gibson2020recycling,ahmad2019hybrid,gharbia2019robotic,cornet2019circular,petrvs2019molemod,rausch2019spatial,sanchez2019multi} & \ref{sec:planning:auto_sp}\\\cline{2-10}
& Mixed-Model Problem Formulation &  &  & $\bullet$ &  &  &  & \cite{ming2019multi,cevikcan2020disassembly,cao2019novel,liu2019many,alvarez2019complexity,rodriguez2019iteratively,murali2019optimal,watson2019assembly,bahubalendruni2019optimal,wen2019research,hui2019research,yang2019design,roldan2019training,li2020efficient,kardos2020constraint,fechtera20193d,li2019disassembly} & \ref{sec_Mixed_Model_Problem_Formulation}\\\cline{2-10}
& Optimization Algorithms &  &  & $\bullet$ &  &  &  & \cite{veenstra2019watchmaker,gunji2019hybridized,suarez2019practical,tian2019product,fang2020multi,lu2020hybrid,fang2019evolutionary,laili2019robotic,xu2020disassembly,tao2019joint,murali2020robotic,yang2019research,wang2019partial,li2019fast,bahubalendruni2019subassembly,tseng2019hybrid} & \ref{sec_Optimization_Algorithms}\\\cline{2-10}
& Optimization Criteria &  &  & $\bullet$ &  &  &  & \cite{pastras2019theoretical,galizia2019product,xia2019balancing,desai2019ease,hu2019charts,marconi2019applying,parsa2019intelligent,guo2019lexicographic,nurimbetov2019robotic,costa2019algorithmic,gregg2019assembled} & \ref{sec_Optimization_Criteria}\\\cline{2-10}
& Evaluation &  &  & $\bullet$ &  &  &  & \cite{li2019multi,malik2019complexity} & \ref{sec_Evaluation}\\\cline{2-10}
& Learning Assembly Sequence Planning &  &  & $\bullet$ &  &  &  & \cite{geft2019robust,franccois2018introduction,zhao2019aspw,zakka2019form2fit,rizwan2019formal,cunha2019towards,papanastasiou2019towards} & \ref{sec_Learning_Assembly_Sequence_Planning}\\\cline{2-10}
& Grasp Planning &  &  & $\bullet$ &  &  &  & \cite{dogar2019multi,ali2020integrated} & \ref{sec_Grasp_Planning}\\\cline{2-10}
& Alleviating the Computational Burden &  &  &  &  &  &  & \cite{gunji2019effect,murali2019integrated,cailhol2019multi,rodriguez2020pattern} & \ref{sec:planning:computational_burden}\\\hline
\multirow{6}{*}{Executing Robotic Assembly and Disassembly} & Types of Robots & $\bullet$ & $\bullet$ & $\bullet$ & $\bullet$ & $\bullet$ & $\bullet$ & \cite{weng2019survey,liu2019inverse,li2019deformation,dragomir2019modelling,zhao2019typical,duque2019trajectory,marconi2019feasibility} & \ref{sec:execution:types_of_robots}\\\cline{2-10}
& Object Grasping and Manipulation &  &  &  &  &  &  & \cite{hietanen2019benchmarking} & \ref{sec:execution:object_grasping_and_manipulation}\\\cline{2-10}
& Compliance &  &  &  &  &  &  & \cite{zhang2019peg} & \ref{sec:execution:compliance}\\\cline{2-10}
& Tackling Uncertainties &  &  &  &  & $\bullet$ & $\bullet$ & \cite{jenett2019relative,hayami2019multi,zhang2019auto} & \ref{sec:execution:tackling_uncertainties}\\\cline{2-10}
& Simulating Execution &  &  &  & $\bullet$ &  &  & \cite{hanson2019industrial,axelsson2019using} & \ref{sec:execution:simulating_execution}\\\cline{2-10}
& Human-Robot Collaboration &  &  &  & $\bullet$ &  &  & \cite{perez2019symbiotic,dianatfar2019task,huang2019case,kongthon2009implementing,ding2019robotic,li2019sequence,huang2019study,askarpour2019formal,pecora2019systemic,liu2020remote,bai2019application,kuang2019one,coupete2019multi,zhang2019six,wang2019dynamic} & \ref{sec:execution:hrc}\\
\hline
\end{tabular}
\end{adjustbox}
\label{19}
\end{table*}

\begin{table*}[t]
\centering
\caption{Some of the most fundamental key performance indicators of a circular economy as suggested in~\cite{li2019multi}.}
\begin{adjustbox}{max width=\textwidth}
\begin{tabular}{|c|c|c|}
\hline
Index & Key performance indicator & Description\\\hline
E1 & Environmental benefit ratio & Environmental impacts arising from material production, recycling or disposal\\\hline
E2 & Target material concentration & Scarce materials and their concentration within the components\\\hline
T1 & Total disassembly time ratio & Ratio of manual to to robotic disassembly time for a given product\\\hline
T2 & Complexity of the robotic disassembly & The number of robotic operations, tool-change operations and joining methods\\\hline
Eco1 & Disassembly cost and revenue ratio & Proportional advantage of the cost and revenue ratio of robotic disassembly with respect to the manual one\\\hline
Eco2 & Total profit from robotic disassembly & Profit from robotic disassembly, i.e. materials' value after subtracting the associated costs\\
\hline
\end{tabular}
\end{adjustbox}
\label{20}
\end{table*}

\begin{figure}[t]
\centering
\includegraphics[width=0.48\textwidth]{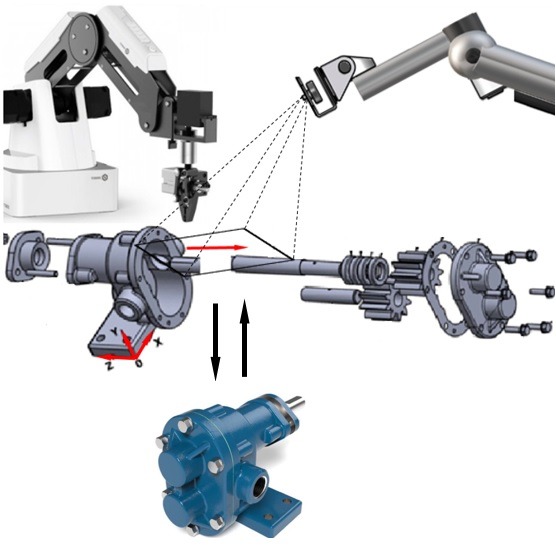}
\caption{A product is assembled from a set of component parts. At the end of its usage it can be disassembled and its components reused or recycled. Robots and artificial intelligence can play a fundamental role in optimizing both, with great benefits of costs, efficiency and environmental sustainability. The figure has been adopted from~\cite{ramirez2020economic}.} 

\label{fig11}
\end{figure}

Automatizing product assembly and disassembly (a process known as \ac[4]{RAD}\footnote{Throughout this paper, RAD will be used to refer to Robotic Assembly and Disassembly, in general. Whenever differences are important, the more specific terms \ac[4]{RA} or \ac[4]{RD} will be used accordingly.} (see Fig.~\ref{fig11}) can lead to enhancements in speed of production and general affordability~\cite{jenett2019relative}.\\
\ac[4]{RAD} has two important parts: \emph{Planning} and \emph{Execution}. \emph{Planning} for \ac[4]{RAD} consists in coming up with a sequence of actions that, given an initial configuration, would result in a desired final one~\cite{ahmad2019hybrid}. This specific type of planning is known as \emph{Sequence Planning}. In the past, \emph{Classical Sequence Planning} used  to  be  done by highly qualified process engineers. A variety of tools like computer-based platforms for creating and storing plans, generative process planning and computer aided design interfaces have considerably automatized the process. Sequence planning is a complex mathematical optimization problem. In the earliest stages, solving it was solely based on mathematical models. Gradually, \ac[4]{AI} came into play, aiming at reducing search spaces and alleviating the resulting computational burden by decreasing the number of iterations required for coming up with an optimal or near-optimal solution. This required rigorous problem formulations, optimisation algorithms, criteria and evaluation and a careful  consideration of computational costs. \\
Upon availability of a plan, a robot would be in charge of \emph{executing} it. There are a variety of available industrial robots each with specific advantages for specific tasks. In order to execute the sequence planning, these robots have to be able to manipulate 3D objects, which requires robust visual perception for pose detection, advanced haptic, tackling uncertainties and errors, and safely and efficiently collaborate with humans.\\
Both planning and execution of \ac[4]{RAD} can be facilitated by product design. Reducing the number of parts, preference for certain geometric shapes, improving accessibility of the robotic arm to parts' attachments are just a few attributes that could contribute to higher efficiency and lower costs. \\
Finally, \ac[4]{RAD} has considerable economic, environmental, and social benefits. It can greatly improve processes in re-manufacturing and recycling, which is fundamental in today's increasingly circular economy.\\
In the following, we offer an overview of \ac[4]{RAD}. Sec. \ref{sec:planning} is dedicated to sequence planning. Sec.~\ref{sec:execution} details key execution aspects. Facilitating planning and execution by design is the topic of Sec. \ref{sec:design_for_assembly}. We continue with a study of the importance of the underlying technologies for sustainability and circular economy in Sec. \ref{sec:circular_economy}. The final section, Sec. \ref{sec:conclusion}, is dedicated to conclusions and recommended future directions of study.\\
Moreover, a summary of the concepts reviewed throughout the paper has been listed in Table~\ref{19}, along with their linkages to some of the most essential circular economy key performance indicators introduced in Table~\ref{20}, as well as the indices of the associated references and relevant sections.

\section{Planning Robotic Assembly and Disassembly}
\label{sec:planning}
\begin{figure*}[t]
	\centering
	\includegraphics[width=\textwidth]{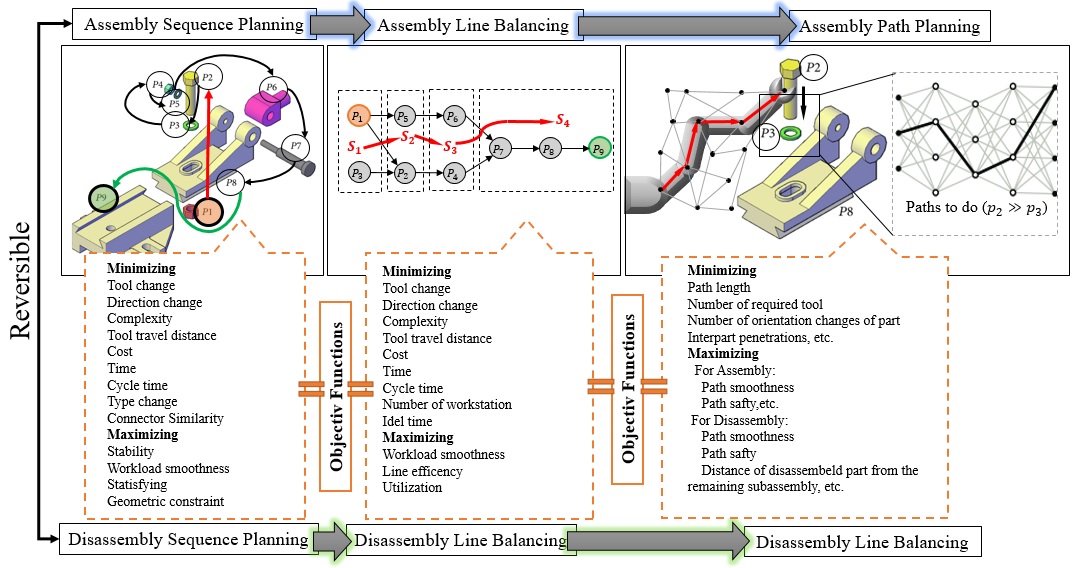}
	\caption{Planning of \ac{RA} and \ac{RD} processes in a glance. The figure has been adopted from~ \cite{ozmen2018optimum}. Common objective functions based on which the main steps of \ac[4]{RA} and \ac[4]{RD} processes may be optimized are listed, where given certain assumptions, each of the foregoing two might be reversible, resulting in the other one. Assembly and disassembly \ac{SP} procedures are aimed at the tasks of devising sequences of collision-free operations resulting in attaching parts to a product being assembled or removing them from the one being disassembled, respectively. Moreover, the line balancing problem is targeted at deciding on the choice and order of the operations to be left upon each of the workstations. Last but not least, the task of path planning concerns finding feasible paths following which each of the operations could be performed by the relevant robot within the corresponding workstation.}
	\label{17}
\end{figure*}

In this section, the most essential elements of planning processes for \ac{RA} and \ac{RD} procedures, which have been summarized in Fig.~\ref{17}, will be reviewed. For more in depth review of \emph{ Assembly Sequence Planning} the interested reader is referred to ~\cite{warren2019computer,abdullah2019optimization}. \emph{Disassembly Sequence Planning} is treated in ~\cite{zhou2019disassembly}, while a survey of studies on \emph{Disassembly Line Balancing Problem} has been provided in~\cite{ozceylan2019disassembly, hazir2019review}.

\subsection{Classical Sequence Planning}
\label{sec:planning:classical_sp}

\ac[4]{RA} is concerned with converting pieces or parts from their original or intermediate forms to the final ones. This typically involves tasks such as machining, forming and joining. For joining different parts of a product, mechanical fastening, adhesive bonding, soldering, brazing and welding are among the most common types of approaches utilized. \\
The order in which these tasks are performed is an important factor for efficiency and speed, resulting in higher quality and lower costs and required amount of raw materials. This affects the product's competitiveness.\\
Classically, it used to be done by a process (or assembly) engineer, who would need to possess a variety of skills, e.g. to be familiar with the approximate costs of the raw materials, tooling procedures and other processes, to understand the role of the raw materials and fixtures, to estimate the time and costs it takes to perform the processes, to be capable of using machinability data handbooks and other references, as well as engineering drawings, and be aware of what resources could be obtained from the shops.\\
Usually, a process planning procedure, which involves \emph{sequence planning} at its core, is performed through iterating over the following steps: 1. Understand the design specifications and the drawings, 2. Upon detecting the datum surfaces, determine the setups, 3. Pick up the suitable machines for each setup, 4. Decide on the sequence of operations for each setup, 5. Choose appropriate tools for each process, 6. Design or opt for the relevant fixtures for every setup, 7. Tune the associated parameters, 8. Make the final process plan. It should be noted that failing to establish reasonable assembly relations leads to various deficiencies, including a waste of the raw materials~\cite{warren2019computer}.

\subsection{Automatizing Sequence Planning}
\label{sec:planning:auto_sp}
A great fraction of early ideas aimed at automatizing process planning were intended to deal with machining processes. It underwent four main phases.\\
First, to develop computer-based platforms making it possible to create, store and retrieve reports and plans.\\
Second, computers were introduced, which could enabled storing plans previously developed, and at the time a new one for a similar product needed to be created, retrieve the foregoing plan, and then revise it according to the associated characteristics.\\
In the third stage, an effort was made to obviate the necessity of the presence of a process engineer, resulting in a generative automatized process planning where the logic based on which decision-making needs to be performed would be handled automatically using some form of \ac[4]{AI}. Ideally it will not demand the use of prior plans.\\
In the fourth phase, Computer Aided Design interfaces were emulated using feature extraction for extracting information from drawings.\\
Nevertheless, despite the great amount of progress made in automatic matching processes, the essential aim of eliminating the role of the human operator and automatizing the whole process planning pipeline did not experience a comparable level of advancement. This was due to a variety of factors, including the particularities of every manufacturing system, the typical enormousness of the process planning knowledge and the deficiencies design models usually entail for process planning~\cite{warren2019computer}.\\
On the other hand, with the recent advances in augmented reality, it is possible to develop semi-automated human-in-the-loop frameworks through employing Computer Aided Design models within robotic simulation software. Given proper calibration, it facilitates the process of programming the assembly of small batch-sized products involving numerous parts in terms of complexity, costs and programming time-consumption associated with practical application~\cite{linnerud2019cad}.\\
Part agent systems, which take advantage of network agents, possibly along with Radio-Frequency Identification and Augmented Reality technologies, enable one to monitor the statuses of parts and products throughout their whole life. They can offer pieces of advice regarding e.g. whether a certain part needs to be changed for new, requires additional care, or is in a form that is good enough for it to be retrieved and used in the remanufacturing of other products. Moreover, such systems could facilitate disassembly processes by storing and visualizing the relevant models~\cite{nagasawa2019disassembly}.\\
Nevertheless, studies show that there is still a wide spectrum of assembly tasks related to the manufacturing of mechanical products, which have not been practically robotized, despite potential possibilities. For example, when it comes to automobiles, only 10 to 20 percent of the relevant operations are performed in an automated manner~\cite{said2019mathematic}. An example of classically manual processes that are being gradually robotized and automated is sorting disassembled parts which is taking more and more advantage of machine learning technologies during the last few years~\cite{gibson2020recycling}.\\
Robots may be able to collaborate for planning the necessary operations and performing the construction tasks required for making structures such as shelters, bridges, buildings or overhang constructions from prefabricated components, such as cylindrical parts, in disaster areas, which are unsafe or hard to access for humans. Some of the challenges involved in task planning for the foregoing purpose arise from the necessity of employing additional components playing the role of counterweights or temporary supports during the operation, preassembling moving structures or integrating prepared structures into the final design.\\
On the other hand, on top of geometric feasibility regarding possible collisions between the blocks, further feasibility checks are required for avoiding collisions while trying to reach a certain block, as well as for guaranteeing stability. In~\cite{ahmad2019hybrid}, answer set programming was utilized in order to develop a hybrid planning system for multiple robots with the capability of performing the aforementioned feasibility checks, including stability and reliability, not only for the final structure, but also for each of the sub-structures in every step of the process of construction.\\
One of the greatest applications of \ac[4]{RA} is in constructing buildings through Rapid Prototyping\footnote{A family of methods to build a scale model of a part or product in a fast manner based on computer-aided design and \ac{3D} printing, or additive manufacturing~\cite{efunda,rp,rp1}.} by means of Robotic Manipulators\footnote{Arm-like mechanisms consisting of series of links and joints referred to as cross-slides, which are utilized for grasping and manipulation without human intervention~\cite{automation,automation1}.}. The first attempts were made in 1980s in Japan, and have, ever since, been vastly studied by researchers from the USA, Germany and Switzerland. Such technologies are practically better applicable to low-rise buildings, and take advantage of recent advances in swarm robotics and 3D printing. Although many of the tasks required for concreting, steel reinforcement and formwork could be left to robots, horizontal Reinforced Concrete elements still cannot be created on-site and in an automated manner~\cite{gharbia2019robotic}.\\
The use of robotics in the construction of buildings takes special importance whenever there is a shortage of skilled workers due to e.g. an aging population, and manifests its advantages as higher speed and lower CO\textsubscript{2} emission~\cite{cornet2019circular}, as well as higher safety and less energy consumption than manual operations~\cite{petrvs2019molemod}. It could accelerate the process by about 5~times, compared to the case where the task is performed manually.\\
\ac[4]{RD} of buildings, including their interior walls, has recently benefited significantly from digitization technologies such as automated meta-data generation and scan-to-Building Information Modelling\footnote{Digital characterization of a place in terms of properties and functionalities, throughout the whole life-cycle, i.e. from formation to demolition~\cite{bmi,bmi1}.}. One of the underlying challenges is to collect and process raw, non-semantic data intelligently, aiming at an automated process of spatial parameterization, which involves extraction, modification and analysis of the spatial properties of a component.\\
Similarly to other applications of \ac[4]{RD}, it will result in adaptive reuse of the building, selective disassembly for repairs and rehabilitation, reconfigurable, prefabricated assemblies and \ac{RA} programming. Using a rule-based algorithm, the 3D Cartesian properties and possible clashes between non-semantic Computer Aided Design elements could be detected. Global spatial constraints for disassembly,  Building Information Modelling, Level Of Development\footnote{The constituents and the level of robustness of different elements of the Building Information Modelling at various stages~\cite{grani_2016}.} and modeling accuracy are among the factors which could affect the resulting level of automation~\cite{rausch2019spatial}.\\
Moreover, selective disassembly planning through Multi-Objective Optimization may lead to combinations of different deconstruction approaches, considering the technical and economic constraints imposed by the underlying components~\cite{sanchez2019multi}.\\
The idea of developing a \ac{CE}, which involves, among others, design for assembly, as one of its main components, is applied to the Amstel~III district in Amsterdam, Netherlands in~\cite{cornet2019circular}. It is meant to develop modular component, which could be connected to each other in a reversible manner. In the aforementioned application, the materials from an office building have been transformed to a mid-rise, mixed-use building that includes residential compartments, as well as a library offering flexible workplaces, an exhibition area and a restaurant.


\subsection{Sequence Planning}
\label{sec:planning:sp}
\begin{figure*}[t]
	\centering
	\includegraphics[width=.8\textwidth]{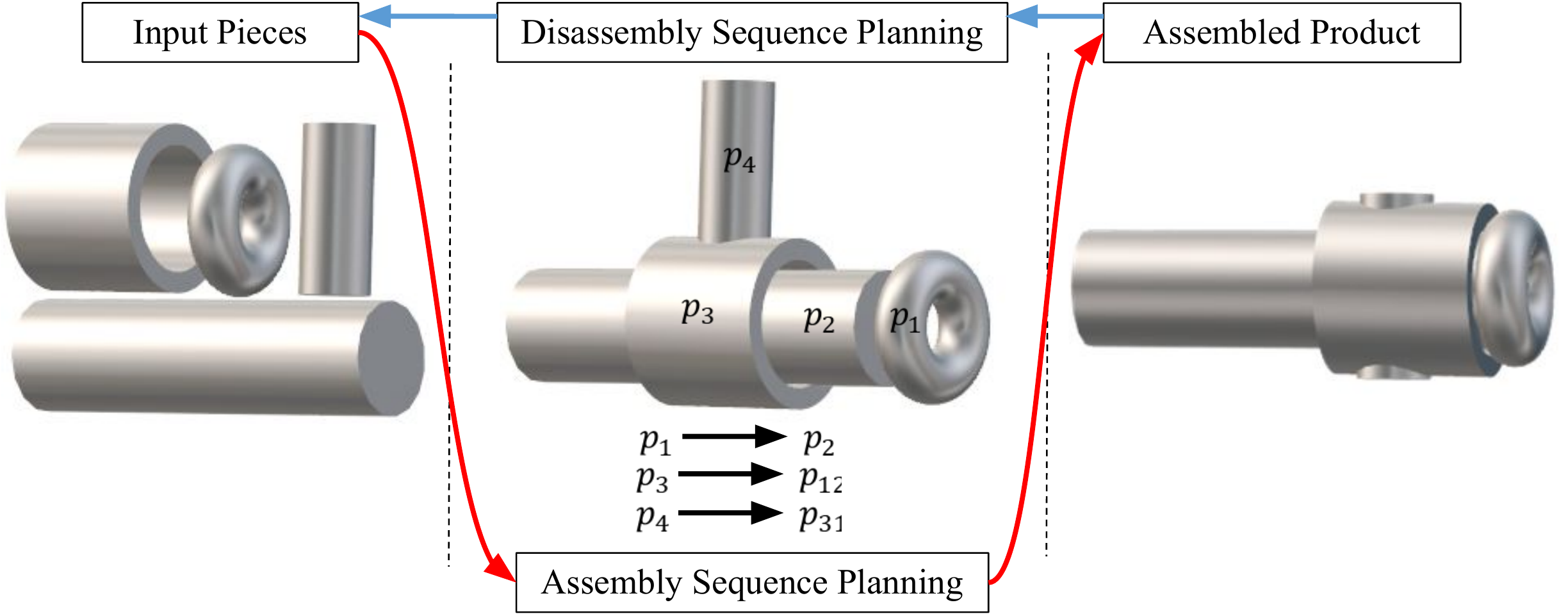}
	\caption{Graphical representation of \ac[4]{RA} and \ac[4]{RD} sequence planning, with the process flow directions being shown using red and blue arrows, respectively. The four pieces $p_i$, $i:\{1,2,3,4\}$, shown in this figure are assembled in the sequence indicated by arrows, where e.g. ${p_1}\longrightarrow{p_2}$ means that $p_1$ is attached to ${p_2}$, resulting in $p_{12}$.}
	\label{18}
\end{figure*}

Generally, as illustrated in Fig.~\ref{18}, the problem of sequence planning could be thought of as \emph{that of planning a sequence of actions which given an initial configuration could obtain a final one}.

In what follows, the main concepts and procedures related to optimizing the processes of \ac[4]{RAD} will be elaborated, where the outcome will be a knowledge model through which, among others, the assembly process, the relevant spatial positions, the product structure and a set of desirable mating features and hierarchical relations between the subassemblies are explained~\cite{peng2019knowledge}.

\subsubsection{Mixed-Model Problem Formulation}
\label{sec_Mixed_Model_Problem_Formulation}
Performing automated industrial \ac{RAD} tasks often requires numerous robots, organised in multi-robot assembly lines, to work together. Robots providing different functionalities may be associated with different energy consumption profiles and process execution times, whose operations need to be optimized for a fast, efficient and reliable performance~\cite{ming2019multi}.\\
Numerous factors could be considered while trying to optimize the performance of a multi-robot assembly line. This could be handled via a \emph{mixed-model problem formulation} employed under an assembly line \emph{balancing problem}, considering the underlying complexity and disassembly modes as well. For example, destructive disassembly modes, regardless of whether the part is detachable or not, are assumed to be associated with high energy and time consumption, and thus, believed to be of low value.\\
Reducing the amount of space required for \ac[4]{RAD} tasks, as well as the lead time, is among parameters which are negatively affected by the level of automation, i.e. employing fewer operators leads to the necessity of larger spaces and longer lead times. Thus the number of human operators could also be considered as an adjustable parameter within the relevant assembly line balancing problem~\cite{cevikcan2020disassembly}.\\
The aforementioned parameters, i.e. the optimization objectives, may involve e.g. CO\textsubscript{2} emission rate, total load density, hazardous task cost, the number of workstations and robots, the amount of resources utilized and the production rate.\\
On the other hand, in the context of disassembly aimed at remanufacturing, the ratio between the output value and the corresponding resources fed as input provides a clear indication of the relevant quality and efficiency~\cite{cao2019novel}. The mixed model could be framed through mixed-integer mathematical programming~\cite{liu2019many}.\\
It is worth noticing that analyzing the complexity of the balancing problem has been long performed based on the perspective proposed in~\cite{gutjahr1964algorithm}, but this has recently been questioned due to the fact that the foregoing study was published before the concept of NP\footnote{Non-deterministic polynomial time.}-hardness was introduced~\cite{alvarez2019complexity}.\\
In essence, \emph{sequence planning is a multi-objective discrete optimization problem}, which needs to be dealt with as an NP-hard combinatorial one, where one needs to find the set of all feasible sequences, in the sense that the outcome has to be collision-free and that the robot has to be able to carry out the task of assembly based on the sequence planned~\cite{rodriguez2019iteratively}.\\
Unfortunately, intractably large search spaces and local optima are among the known problems arising while trying to handle the foregoing task. To remedy this, stability graphs could be utilized for coming up with stable assembly subsets, which reduces the search space to a smaller one. Moreover, the fitness of the subassemblies could be analyzed at each stage, according to the user's preferences, so as to further reduce the search space~\cite{murali2019optimal}.\\
One of the most crucial criteria in this regard is the level of freedom in movement which is provided to the robot or the cooperating human being. An essential constituent of such an analysis is a Disassembly Interference Graph, which describes the obstruction relations between the parts. It can contribute to performing sequence planning such that the assemblies are partitioned with access to the parts taken into account as a major priority, i.e. considering minimizing blockage of the parts as the target. Consequently, a series of subassembly decomposition is developed in the form of a tree structure that is suitable for parallelization~\cite{watson2019assembly}. It is worth noticing that parallelization of assembly or disassembly tasks requires ensuring the stability of the subassemblies, which demands inferring and taking into account extended subassembly stability relationships, aiming at avoiding possible undesired divergences of the main product structure during the course of performing \ac[4]{RAD}~\cite{bahubalendruni2019optimal}.\\
One of the fundamental topics in the context of sequence planning is generating assembly precedence constraints, which could be performed based on 3D models in a simulation environment, using block sequences, which enhances the quality, and reduces the costs associated with developing Virtual Assembly software. Virtual assembly of a four-coordinated mobile robot has been reported in~\cite{wen2019research}. Design and assembly of marine diesel engines is another instance of the applications of the technologies discussed throughout this article, which could be performed through visual 3D simulation~\cite{hui2019research}. The development of a robotic reconstruction site has been reported in~\cite{yang2019design}, where sensors have been utilized for simulating real-world data through a virtual environment. Learning to assemble a toy with building blocks from demonstrations by expert users through Virtual Reality and process mining has been presented in~\cite{roldan2019training}.\\
In the above context, the block sequence structure is meant to store and process the reference variables of the paths and the attributes resembling assembly precedence constraints. In order to bypass the requirement of preprocessing each part model, assembly reference objects are considered as nodes for distributed management of the aforementioned paths. This way, preprocessing and calculation are performed on the precedence relations rather than a particular assembly or disassembly sequence. Such a software could be developed together with the relevant state-transition functionalities using the procedures introduced in~\cite{li2020efficient}, where the required parameter-input interfaces are available for a variety of simulation platforms.\\
The main challenges involved in computer-aided process planning arise from the economical, technological and geometrical constraints. Decomposing the problem into smaller ones, including macro-level problem targeting and managing the general process and micro-level ones concentrating on the details of individual assembly or disassembly operations, may contribute to reducing the foregoing complexity. To this end, a constraint model could be used, where assembly operations, as well as the whole product, are resembled through feature-based representations, which constitute the macro-level problem. On the other hand, feedback from micro-level, which is represented as feasibility cuts, helps integrate the solutions to the problems into a single, efficient computer-aided process planning package~\cite{kardos2020constraint}.\\
Traditionally, an experienced design engineer would need to extract the properties of the suitable \ac[4]{RAD} process of the product, as well as its own features, from the associated Computer Aided Design models. However, the foregoing task has greatly advanced during the last few decades, and has gradually become more automatic within typical industrial settings, which is based on product liaison matrices based on symmetric feature detection, and results in a standard pose and orientation exchange file format~\cite{fechtera20193d}.\\
On the other hand, at the design stage, the trajectory following which each part needs to be taken from the original pose and orientation to the one at which it could be mounted in a collision-free manner has to be determined in advance. Achieving the foregoing objective may require creating a multi-directional combined trajectory, which could be realized through employing the bounding box approach and the notion of position matrix~\cite{said2019mathematic}.\\
Furthermore, the equipment required for an assembly or disassembly line constitute a significant factor in estimating and comparing different alternatives of doing so. Thus the choice and configuration of the equipment could be modeled and analyzed in conjunction with the main assembly line balancing problem, in order to enhance the chances of succeeding in reducing the overall cost associated with the design of an assembly/disassembly line, which could be formulated as a Mixed Integer Linear Programming problem~\cite{li2019disassembly}. In this context, the latter refers to an optimization program some of whose variables are bound to take only integer values, where the objective function and the constraints are linear~\cite{milp,integer_programming}. According to~\cite{benichou1971experiments}, assuming that $\mathbf{X}$ and $\mathbf{Y}$ are column vectors with the components $x_i$, $i=1$ to $p$, and $y_j$, $j=1$ to $q$, which stand for the continuous and and integer variables, respectively, and $\mathbf{A}$ and $\mathbf{B}$ are rectangular matrices of sizes $(m,p)$ and $(m,q)$, respectively a Mixed Integer Linear Programming may be mathematically formulated as follows:
\begin{equation}
\begin{aligned}
\min_{\mathbf{X},\mathbf{Y}}\quad&\mathbf{F}=\mathbf{A}_0^\text{T}\mathbf{X}+\mathbf{B}_0^\text{T}\mathbf{Y},\\
\textrm{s.t.}\quad&\mathbf{A}\mathbf{X}+\mathbf{B}\mathbf{Y}=\mathbf{D},\\
&\alpha_j\leq y_j\leq\beta_j,\\
&y_j\text{ integer},&j=1\text{ to }q,\\
&0\leq x_i,&i=1\text{ to }p.
\end{aligned}
\end{equation}
which can be solved for an optimal solution, possibly among several of them, using Dynamic Programming. The latter refers to a family of useful tabular algorithms for solving optimization problems involving overlapping subproblems, i.e. sharing subsubproblems. In order to solve the main problem, Dynamic Programming resorts to combining solutions to the subproblems. To avoid computing solutions to the subsubproblems recurrently, it stores them into a table.\\
Typically, a Dynamic Programming algorithm comprises of, first, identifying the problem's structure, and then defining the value of an optimal solution recursively, being pursued by computing the value of the optimal solution in the aforementioned bottom-up fashion. If needed, the corresponding solution itself may also be found based on the resulting information, being, in some cases, supplemented by additional data kept for a more efficient computation~\cite{cormen2009introduction}.

\subsubsection{Optimization Algorithms}
\label{sec_Optimization_Algorithms}
Generally, optimizing the design and performance of robotic systems is inspired by evolutionary processes and their dynamics. The foregoing approaches may be categorized based on whether they try to develop a new solution or aim at improving upon the ones that already exist. Nevertheless, typically, most of them prefer to find a compromise between the two, i.e. exploration and exploitation, through incorporating intrinsic mortality, which is connected with the mutation rate as well. Similarly, a trade-off needs to be found between genotype and phenotype mapping, referring to the direct or generative manner of expressing the genomic information, respectively.\\
Lindenmayer systems (L-systems) could be used for developing a generative encoding enabling to make virtual creatures inspired by plants. In the particular case of modular robots, generative encoding have played a major role in development by enabling mapping the genotype to the phenotype. It has been shown to result in better locomotion performances compared to direct encoding. For example, adding extra solar panel modules for energy autonomy could provide inspiration for the evolution of various types of robots, an example being knife-fish-inspired soft robots~\cite{veenstra2019watchmaker}.\\
Recent advances in soft computing have contributed to improving the efficiency of Optimal Assembly Sequence Planning algorithms in terms of CPU times and fitness values~\cite{gunji2019hybridized}. Nevertheless, typically, the greatest challenges involved in finding out an optimal or near-optimal solution in the context of assembly sequence planning arise from intractable largeness of the search spaces, which results in longer execution procedures, and the possibility of getting stuck at local optima or dead ends, greatly affecting the performance and robustness of methods such as  Labeled Real-Time Dynamic Programming, policy iteration or value iteration~\cite{suarez2019practical}.\\
Disassembly line balancing problems, including the ones involving multiple products with different models, are NP-hard, as a special case of the bin packing problem~\cite{alvarez2019complexity}, and could be solved using mixed-integer mathematical programming~\cite{ming2019multi}, followed by utilizing a variant of genetic algorithms, which makes use of chromosome evolution rules, e.g. operators performing mutation, crossover or selection~\cite{tian2019product}. For example, the Many-objective Best-order-sort Genetic Algorithm, which involves its own neighborhood search operator, two genetic operators, encoding/decoding scheme and best-order-sort mechanism, could be utilized~\cite{liu2019many}. An alternative approach could be to apply meta-heuristic simulated annealing evolutionary optimization~\cite{fang2020multi}.\\
When optimizing the performance of multi-robot disassembly lines, it should be duly taken into account that different robots may be associated with different energy consumption rates and task times. On the other hand, in the cases of uncertainties in task processing times, intervals representing upper and lower bounds on them, which may be known from experimental and physical conditions, could be considered instead~\cite{fang2020multi}.\\
Being motivated by environmental sustainability concepts, the tasks of assembly sequence planning needs to be performed such that the profitability and energy consumption could be taken into account as concurrent optimization criteria, which gives rise to problems referred to as  Profit-oriented   and Energy-efficient Assembly Sequence Planning~\cite{lu2020hybrid}.\\
On the other hand, multi-objective evolutionary optimization could be considered for solving a mixed-model balancing problem for multi-robot workstations. Among other factors that affect the performance such frameworks, the manner of initialization and variation operator taken into account could be mentioned, which may be devised in a problem-dependent fashion~\cite{fang2019evolutionary}.\\
Ternary bees algorithms constitute another category of the optimization techniques which could be utilized in this context. The virtues of meta-heuristic methods and greedy search could be resorted to simultaneously, based on concurrent operations and collaborative potential solutions~\cite{laili2019robotic}. Another relevant example is the dual-population discrete artificial Bee Colony algorithm~\cite{cao2019novel}.\\
The Disassembly Sequence Planning problem could also be handled using, for instance, the  Modified Discrete Bees Algorithm based on Pareto~\cite{xu2020disassembly}. Nevertheless, it should be noted that for scheme selection and recovery route assignment, once obtained, Pareto-optimal sets of solutions resulted from hybrid meta-heuristic multi-objective optimization algorithms that are based on symbiotic evolutionary mechanisms need to be further studied through decision-making processes such as the fuzzy set method, in order to pick up the best possible solutions, which may rely on the preference factors supplied by the user or operator, and aims at picking up options which are, at least, non-inferior to the rest of the solutions~\cite{tao2019joint}.\\
The improved fruit fly algorithm is also an alternative to the problems of optimal Assembly Sequence Planning~\cite{murali2020robotic}. Other studies have reported successful application of Petri Net, as well~\cite{yang2019research}. The flower pollination algorithm, which employs discrete operations and four heuristic rules, is another option for solving such multi-object optimization problems~\cite{wang2019partial}\\
Finding an optimal or near optimal solution from the search space could also be performed using  Markov  Decision  Processes, where heuristic search and determination based on on-line resolution help achieve higher robustness in comparison to methods such as Labeled Real-Time/Near Real-Time Dynamic Programming and optimal control. In the foregoing settings, hierarchical sub-tasks of a simpler and smaller nature, which may be structured automatically, improve the feasibility and reliability of the relevant procedures. More clearly, decision-making for probabilistic planning could be performed according to stochastic shortest path problems being dealt with using Markov Decision Processes~\cite{suarez2019practical}.\\
Disassembly line balancing problems could be solved using classical software packages, such as IBM ILOG CPLEX Optimization Studio~\cite{manual1987ibm},
or alternatively, a branch, bound and remember algorithm with AND/OR precedence. The memory-based dominance rule could be utilized for removing duplicate sub-problems, where cyclic Best-First  Search could be resorted to in order to accelerate the process of coming up with high-quality, complete solutions~\cite{li2019fast}.\\
In~\cite{bahubalendruni2019subassembly}, the Elephant Search Algorithm has been used for optimizing the number of \ac[4]{RA} levels in the context of Optimal Assembly Sequence Planning.\\
Hybrid  Cuckoo–Bat  Algorithm,  Ant  Colony  Optimization, Grey Wolf Optimization, Advanced   Immune System   and   Hybrid   Ant–Wolf Algorithm are other instances of optimization approaches that have been applied to the problems of Optimal Assembly Sequence Planning in the literature~\cite{gunji2019hybridized}, along with Bidirectional Ant Colony Optimization as an alternative approach with a relatively higher efficiency than the aforementioned algorithms~\cite{tseng2019hybrid}.

\subsubsection{Optimization Criteria}
\label{sec_Optimization_Criteria}
\begin{figure}[t]
	\centering
	\includegraphics[width=.48\textwidth]{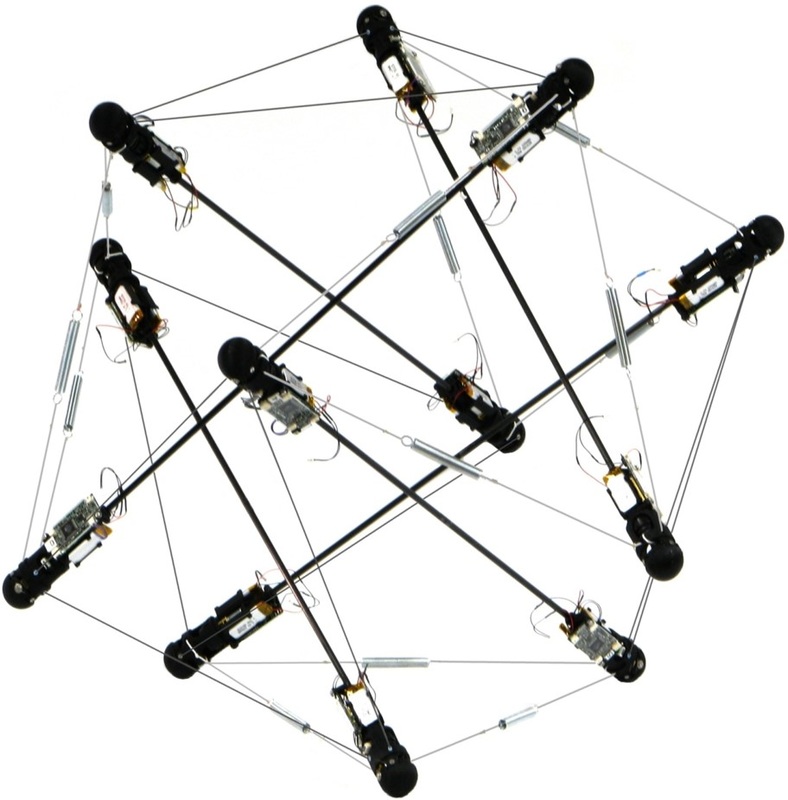}
	\caption{A Tensegrity robot. Figure from~\cite{creative_machines}.}
	\label{16}
\end{figure}

A \acp[4]{DLBP} is aiming at allocating resources to processes in an optimal manner~\cite{fechtera20193d}. Numerous types of criteria or combinations of them could be considered when devising and solving such a problem. Common choices include minimizing total and peak workstation energy consumption, as well as cycle time and the number of robots operating simultaneously~\cite{fang2019evolutionary}. More clearly the disassembly's financial benefits, i.e. the recovery profile, and the consequent environmental and economic impacts, and energy consumption are considered conflicting objectives. Finding a balance between them needs to be investigated based on a set of prescribed preferences~\cite{tao2019joint}.\\
It is possible to reduce the energy consumption associated with the manufacturing of a certain product by optimizing the acceleration profile, without sacrificing the speed and convenience, i.e. with the same production cycle and equipment~\cite{pastras2019theoretical}.\\
Moreover, if applicable, the cost associated with hazardous tasks could be considered as an extra objective~\cite{ming2019multi}. In contexts such as the assembly of re-configurable structures, other common optimization criteria include fault tolerance and the total distance travelled~\cite{costa2019algorithmic}.\\
Other factors involved in typical \acp[4]{DLBP} include the consistency in the amount of idle time that  different workstations experience, as well as fast or prior manual removal of highly demanded or dangerous parts.\\
On the other hand, a change of the tool or its current operational direction could also be considered as costs in the contexts of Optimal Assembly Sequence Planning. Similarly, the disassembly feasibility of each part could be taken into account for choosing the operations that e.g. are easiest, or result in the disassembly of the most wanted parts. This has given rise to indices formulating the operational time and cost, which will be introduced in what follows.\\
Although most of the relevant studies reported in the literature heretofore have considered the time and cost associated with the \ac[4]{RA} procedure as the main criteria for Optimal Assembly Sequence Planning, a number of additional factors may result in the outcomes of the foregoing calculations being rendered unrealistic or inaccurate. First of all, despite the prevalent assumption that the product needs to be disassembled completely, this might not be the case under various scenarios, which implies that a partial disassembly of the product, depending on the underlying purpose, may need to be considered and evaluated. Second, the time and cost may depend largely on the \ac[4]{AS} taken into account, where some of the existing indices may fail to provide a sound, reliable and effective evaluation and comparison of all possible orders in which the operations are thought to be performed. Third, End-of-Life\footnote{A product at the end of its useful life.} products of the same type may be in different conditions at the stage of \ac[4]{RD}, which may also affect the extent to which the results of the above analyses corresponds to the reality.\\
To address the above shortcomings, on top of the \emph{Disassembly Cost Index}, which assesses the cost and time of the \ac[4]{RD} procedure, indices such as Disassembly Handling Index and Disassembly Operation Index have been introduced, aiming at evaluating the feasibility and and complexity of the relevant operations. They result in easier and faster decision-making through obviating the necessity of actually disassembling the \ac[4]{EoL} completely, as well as making it possible to discriminating between the efficiencies of different \ac[4]{RD} procedures aimed at \ac[4]{EoL} products of the same type being in non-identical conditions. Furthermore, the \emph{Disassembly Demand Index} has been suggested for finding out the priorities of different components being disassembled from an \ac[4]{EoL}, based on the demand for each of them~\cite{parsa2019intelligent}.\\
A \emph{Decision Support System} utilized in the context of \ac[4]{RAD} is aimed at finding a compromise between the number of tasks and the platforms' variety, as well as customizing the selected platforms for the target product variant. Fast, high-variety, customized manufacturing, from a broad perspective, requires platform design. For example, \emph{Median-Joining  Phylogenetic Networks}, which have classically been utilized in biology for predicting living species' ancestry based on the gaining or losing of the genes~\cite{bandelt1999median}, and later adopted in this settings in~\cite{hanafy2015developing}, may help determine the number and composition of the platforms. Providing a chance to consider a flexible number of products, by means of phylogenetic tree decomposition, which necessitates establishing product family phylogenetic networks, different variants of a product may similarly be achieved through gaining or losing of the components.\\
Consequently, trade-offs need to be found, at each level, between the number of \ac[4]{RAD} tasks required for producing a customized variant of the product and the number of platforms, which represents the extent of variety, implying reductions in the delivery times and costs. Such analyses and comparisons could be performed using indices such as \emph{Platforms Reconfiguration  Index}, which represents the effort required for changing the configuration to a different variant, based on the number of assembly and disassembly tasks involved, as well as \emph{ Platforms Customisation Index}, which represents the ease of the assembly and disassembly factors. The results may also be useful in terms of making a decision as to whether to consider Delay Product Differentiation or Assemble to Order strategies for a specific product variant~\cite{galizia2019product}.\\
Other optimization criteria involved in mixed-model problem formulations for \ac[4]{OASP} and \ac[4]{ODSP} include load balancing indices and frequency and costs of invalid operations~\cite{xia2019balancing}.\\
Moreover, obviously, the speed and rate at which products are assembled or disassembled plays a major role in the overall success and efficiency of the associated pipeline. This, in turn, depends on a variety of factors, including the type and number of fasteners, the type of the material from which the product is made and its weight, geometric shape and center of gravity. In fact, such analyses take importance even in the stage where the product has not been designed yet, since the outcomes will help more quickly come up with a suitable assembly strategy. In the foregoing context, design attributes and features are converted to standard time data in the form of Time Measurements Units, using  Time Measurements Methods~\cite{desai2019ease}.\\
On the other hand, estimating the time \ac[4]{RD} will take to be performed on e.g. an \ac[4]{EoL} product is required for conducting decision-making. At the detail stage, it may be estimated using the Boothroyd and Dewhurst classification approach, which provides a framework for systematically representing various features associated with each part, playing a role in operations such as fastening, insertion, orientation, movement and acquisition, as well as other constituents of the procedure which do not concern a certain part, such as terminating the \ac[4]{RAD} procedure~\cite{boothroyd_dewhurst_2011}.
Acquisition, manual processing and insertion constitute typical operations and the time each of which will take needs to be estimated. Since component symmetry features are usually irrelevant in the context of \ac[4]{RD}, the remaining procedures whose required times need to be estimated involve part-removal and replacing time, as well as other environmental variables such as accessibility and visibility, along with all other problem-specific factors. Experiments involving 31 undergraduate students revealed that the difference between the time estimated based on the strategy described above and the time it took to manually perform assembly and disassembly had differences as small as 7.4\% and 2.4\%, respectively~\cite{hu2019charts}.\\
Different properties of the parts involved in each product need to be taken into account, such that the \ac[4]{RA} time could be estimated reliably. Wear, rust and deformation are some of such characteristics, which could be represented by corrective factors that are derived using a data mining procedure based on a relevant demonstration of disassembly. For example, this has been performed on the water pump of a coffee machine and the electric motor of a washing machine with maximum deviations of 3\% and 6\%, respectively~\cite{marconi2019applying}.\\
It should be noted that the \ac[4]{RD} time associated with different \ac[4]{EoL} items of the same product may not all correlate precisely, due to the partially unpredictable statuses of \ac[4]{EoL} products~\cite{parsa2019intelligent}.\\
Finally, the problems of \ac[4]{OASP} or \ac[4]{ODSP} may be subject to various constrains stemming from e.g. the precedence constraints or the resources, such as tools or operators, being limited~\cite{guo2019lexicographic}.\\
In order to achieve lighter weight and greater flexibility by means of incorporating tensile elements, the stability of each component of a structure meant to be used in zero-gravity conditions may be arranged based on a balance between the compressing forces pushing on it and the tensile ones pulling on it, which would appear to try to shorten or extend it, respectively. Tensional Integrity (shortened \emph{Tensegrity}) takes advantage of a continuous network of tension components, where the compression components are discontinuous, in contrast to those of typical man-made structures~\cite{pugh1976introduction}.\\
Tensegrity structures (illustrated in Fig.~\ref{16}) could be utilized for various outer-space applications. They can be assembled using sampling-based motion planning, where forward elimination search is used for estimating the pose and orientation of the structure with respect to the robot. Subsequently, a feasible assembly sequence can be obtained using backward disassembly search, followed by creating motion plans for attaching strings to the bars using rapidly-exploring Random Trees, in order to come up with the desired configuration~\cite{nurimbetov2019robotic}.\\
Assembling large-scale reconfigurable structures constitutes a significant application of robotic assembly. It is important in launching applications that could not be realized instantaneously. Its benefits arise from the capability of assembling modular structures, which make it possible to traverse and revise the configuration. In fact, in the foregoing context, the problem is defined as that of transforming a structure from the original configuration to a goal one~\cite{costa2019algorithmic}.\\
The scope and practicability of outer-space missions could be greatly enhanced through utilizing on-orbit robotics and assembly, whereby, outer-space exploration hardware could be reused for a lower cost. Studies such as~\cite{gregg2019assembled} have shown that the assembly and performance efficiency may be kept roughly constant while incorporating infinitely scalable reconfigurability by employing modular materials, e.g. reversibly assembled cellular composites. Moreover, they have concluded that discretely assembled materials and reversible mechanical connection hardware lead to identifiable, bounded and tolerable parasitic mass penalties. On the other hand, the energy consumption associated with on-orbit tasks, such as conventional forming procedures, additive manufacturing through deposition, recycling, reuse and reconfiguration, could be significantly alleviated by means of taking advantage of systems comprising of re-configurable materials.

\subsubsection{Evaluation}
\label{sec_Evaluation}
As aforementioned the Multi-Objective \ac[4]{DLBP} needs to be solved keeping in mind that a balance is required between multiple criteria, including economic reasonability, technological efficiency and feasibility, favorability from the point of view of environmental sustainability. The quality of the underlying trade-off may be assessed using the Decision Support System proposed in~\cite{li2019multi}.\\
On the other hand, the quality of an \ac[4]{AS} or \ac[4]{DS} could be evaluated based on the underlying complexity, whose importance arises from its effect on the extent to which automation could be realized. For example, the complexity of tasks such as part feeding, mounting and handling, as well as implementing provisions related to human safety, are affected by the aforementioned complexity, and influence the time required for deployment or changeover processes. The complexity depends on the task descriptions and the parts' physical attributes~\cite{malik2019complexity}.

\subsection{Learning Assembly Sequence Planning}
\label{sec_Learning_Assembly_Sequence_Planning}
Two-dimensional geometric planning for e.g. operations consisting of one-step linear translations could be performed using Convolutional Neural Networks~(CNN)\footnote{A special type of neural networks which have convolutions instead of matrix multiplications within at least one of their layers~\cite{goodfellow2016deep,cnn}.}, where a robustness score, i.e. success probability, is attributed to each assembly motion by adding stochastic noise in a simulation environment. In fact, utilizing CNNs obviates the necessity of performing simulation for all of the planned motions, which takes an order of magnitude longer, by mapping each motion to a robustness score, based on a pair of images of the subassemblies before and after they are mated~\cite{geft2019robust}.\\
Intractable expansion of assembly relations and sparse rewards are among the known challenges associated with the use of Deep Reinforcement Learning\footnote{A machine learning strategy taking simultaneous advantage of reinforcement learning and deep learning, where the former concerns the ability of a computational agent to learn how to make decisions through trial and error, i.e. to figure out what to do in order to optimize an objective function, and the latter helps it to do so based on large unstructured input data, obviating the necessity of manual manipulation of the state space~\cite{franccois2018introduction,drl}.} for \ac[4]{ASP}, which may be alleviated through employing parameter transfer and curriculum learning~\cite{zhao2019aspw}.\\
For example, the application of human-robot interaction in \ac{RAD} is demonstrated by the the Form2Fit system, which involves robot learning to assemble a kit from a demonstration of its disassembly. This has led to an average performance rate of 90\% under various initial conditions on kit and object pose and orientation, 94\% under new configurations of the same kits, and 86\% in the case of unfamiliar objects~\cite{zakka2019form2fit}.\\
Moreover, a furniture assembly system has been introduced in~\cite{rizwan2019formal}, where the practical assembly of a table has been presented as a case-study. Similarly, performing construction tasks based on demonstrations by multiple tutors has been implemented and reported~\cite{cunha2019towards}. Last but not least, case-studies on the \ac[4]{RA} of white goods through \ac[4]{HRC} have also been reported in~\cite{papanastasiou2019towards}.

\subsection{Grasp Planning}
\label{sec_Grasp_Planning}
When it comes to performing the planned sequence of \ac[4]{RAD} tasks, one needs to determine the robot configuration through which it could grasp the relevant part and carry out the rest of the necessary transfers while ensuring the feasibility of the planned actions in terms of e.g. collision-avoidance through performing a search strategy over a connected constraint graph aiming at solving a Constraint Satisfaction Problem. The shape the foregoing problem takes is affected by whether or not Human-Robot Collaboration is involved, as well as on the number of robots involved~\cite{dogar2019multi}.\\
\emph{Grasp planning} aims at finding the optimal posture of a robotic hand or gripper, as well as the relevant contact placement, while grasping an object or a part~\cite{lin2015grasp}. \\
Assuming that multiple grasp positions and orientations may be available, one part of the task of planning needs to be dedicated to picking up the most suitable ones depending on the initial and final poses, as well as the rest of the environmental variables, including the part of the product which has already been assembled.\\
An orientation graph may be created which under the absence of feasible poses from the initial to the final poses could generate a series of intermediate poses which could approach the foregoing aim while avoiding collisions. Moreover, in the case of disconnections from the tree, which may happen to a node and its descendants due to the collisions taking place because of the changes in the task space over each step of assembly, the connection needs to be reestablished for further extension of the tree. This could be realized using algorithms such as the two-stage extended  Rapidly-exploring  Random Trees~\cite{ali2020integrated}.

\subsection{Alleviating the Computational Burden}
\label{sec:planning:computational_burden}
Although the computational load associated with \ac{RAD} tasks may be large, various strategies have been proposed to enhance the feasibility of the relevant systems.\\
One way is to incorporate geometric reasoning and planning into the pipeline handling symbolic task planning and assembly planning. This way, the high-dimensional nature of the problem resulting from e.g. the high number of Degrees-of-Freedom\footnote{The number of independent parameters based on which the configuration or state of a mechanical system may be defined~\cite{hale1999principles,dof}.} of the robot or from the complexity of the underlying tasks will be accounted for.\\
It should be noted that in the presence of considerations related to secondary parts, such as fits, washers, bolts and nuts, one may notice a considerable impact of whether or not taking them into account as primary ones on the computation time~\cite{gunji2019effect}.\\
When it comes to \ac[4]{ASP} for new variants of a given product, in order to accelerate the process of sequence planning, it might be possible to develop a system that could perform the foregoing task automatically based on the product specifications, thereby obviating the necessity of the time-consuming and labor-intensive task of doing it manually. Nevertheless, checking for collisions requires performing simulations, which usually slows down the whole process.\\
Alleviating the aforementioned problem could be realized through trying to avoid unnecessary slow checks in the cases where it is possible to find out a violation of the feasibility requirements from faster checks performed at lower levels of fidelity. This could be achieved by iteratively revising feasibility checks at different levels, i.e. from a level considering parts only, which could be performed relatively fast, to high-fidelity ones such as motion planning or full robotic simulations, which are comparatively slow. Moreover, propagating the errors to the higher levels as symbolic rules, they could be taken into account for further reducing the search space, i.e. pruning the search tree~\cite{rodriguez2019iteratively}.\\
Parallel assembly helps distribute the tasks to multiple processes, thereby reducing the required time and enhancing the efficiency associated with the \ac[4]{RAD} of products consisting of multiple sub-assemblies. In this context, the number of assembly levels directly affects the assembly cost and time. To this end, parallel assembly possibilities need to be detected through sub-assembly identification techniques~\cite{bahubalendruni2019subassembly}.\\
Numerous studies have attempted to reduce the computational time required for the complex \ac[4]{ASP} or \ac[4]{DSP} of products consisting of a large number of parts. Examples include the linear-weight Scatter Search algorithm for Multi-Objective Optimization or the lexicographic Scatter Search approach introduced in~\cite{guo2019lexicographic}.\\
Although numerous alternatives for solving an \ac[4]{OASP} exist, the product design itself plays a major role in both elevating the complexity of the problem and the chances of arriving at desirable solution. Therefore, the concept of Design for Assembly, being combined with the \ac[4]{OASP}, may contribute to the feasibility and success of the planning \ac[4]{RA} through e.g. reducing the number of parts. The integrated problem could be solved using optimization methods such as Ant Colony Optimization, considering directional changes as the fitness function~\cite{murali2019integrated}.\\
On the other hand, the computational requirements associated with Constraint  Satisfaction Problems dealing with grasp planning may quickly become intractable when number of operations increase. The foregoing issue may be alleviated for achieving a faster performance than that of directly solving the connected constraint graph through dividing the Constraint Satisfaction Problems into a set of independent smaller problems. More clearly, the set of feasible re-grasps over which the search needs to be performed could be reduced by focusing on local possibilities and removing an increasing number of re-grasps per iteration. Nevertheless, it should be duly taken into account that under certain circumstances, the cost related to the elimination of the re-grasps may be even larger than executing them. In such cases, the process of removing the re-grasps could be stopped, followed by executing the existing best plans~\cite{dogar2019multi}.\\
When it comes to trajectory planning, purely geometrical models may become computationally costly for severely constrained environments or complex products. It has been shown that incorporating higher-level abstractions involving multi-layer environment models containing semantic and topological aspects within a global planner may alleviate the foregoing issue, where the initial coarse model is locally refined based on semantically characterized geometry~\cite{cailhol2019multi}.\\
On the other hand, knowledge transfer frameworks enable platforms allow already acquired information regarding e.g. constraints to the new one using which a newly customized product needs to be robotically assembled or disassembled. In fact, the advantage of the foregoing strategy arises from the possibility of restricting the \ac[4]{RAD} search space based on the prescribed constraints, thereby reducing the amount of required computations. Algorithmic implementation of such an approach involves deducing semantic assembly constraints based on feasibility, performing semantic feature-based graph matching over the assemblies, and finally, employing classification and pattern recognition in order to map the existing knowledge of subassembly constraints to full assemblies~\cite{rodriguez2020pattern}.

\section{Executing Robotic Assembly and Disassembly}
\label{sec:execution}
A review of systems aimed at assisting the operator with the task of assembly, such as \ac[4]{HRC} and context-aware instructive frameworks, has been reported in~\cite{wolfartsberger2019perspectives}. For a review of \emph{\ac[4]{HRC} systems for disassembly} the reader is referred to~\cite{liu2019human}. \\
\subsection{Types of Robots}
\label{sec:execution:types_of_robots}
\begin{figure*}[t]
	\centering
	\subfigure[]{\includegraphics[height=0.16\textheight]{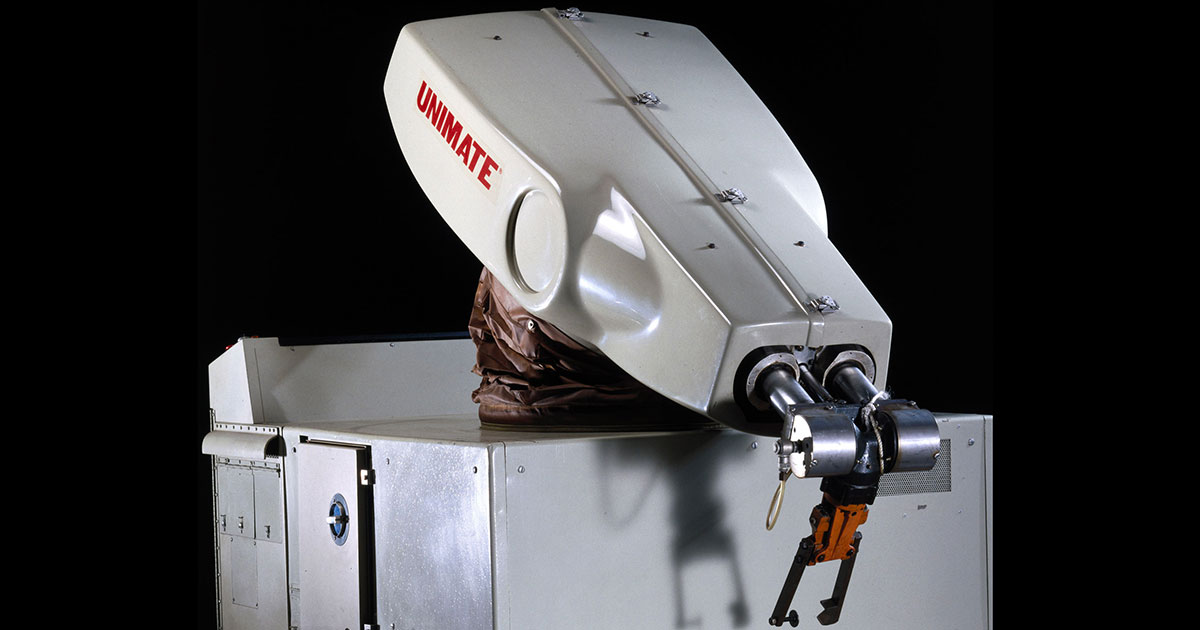}\label{15-4}}
	\subfigure[]{\includegraphics[height=0.16\textheight]{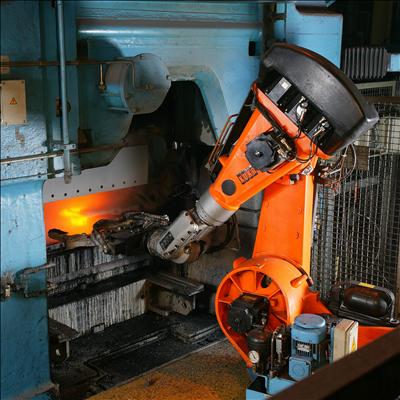}\label{15-1}}
	\subfigure[]{\includegraphics[height=0.16\textheight]{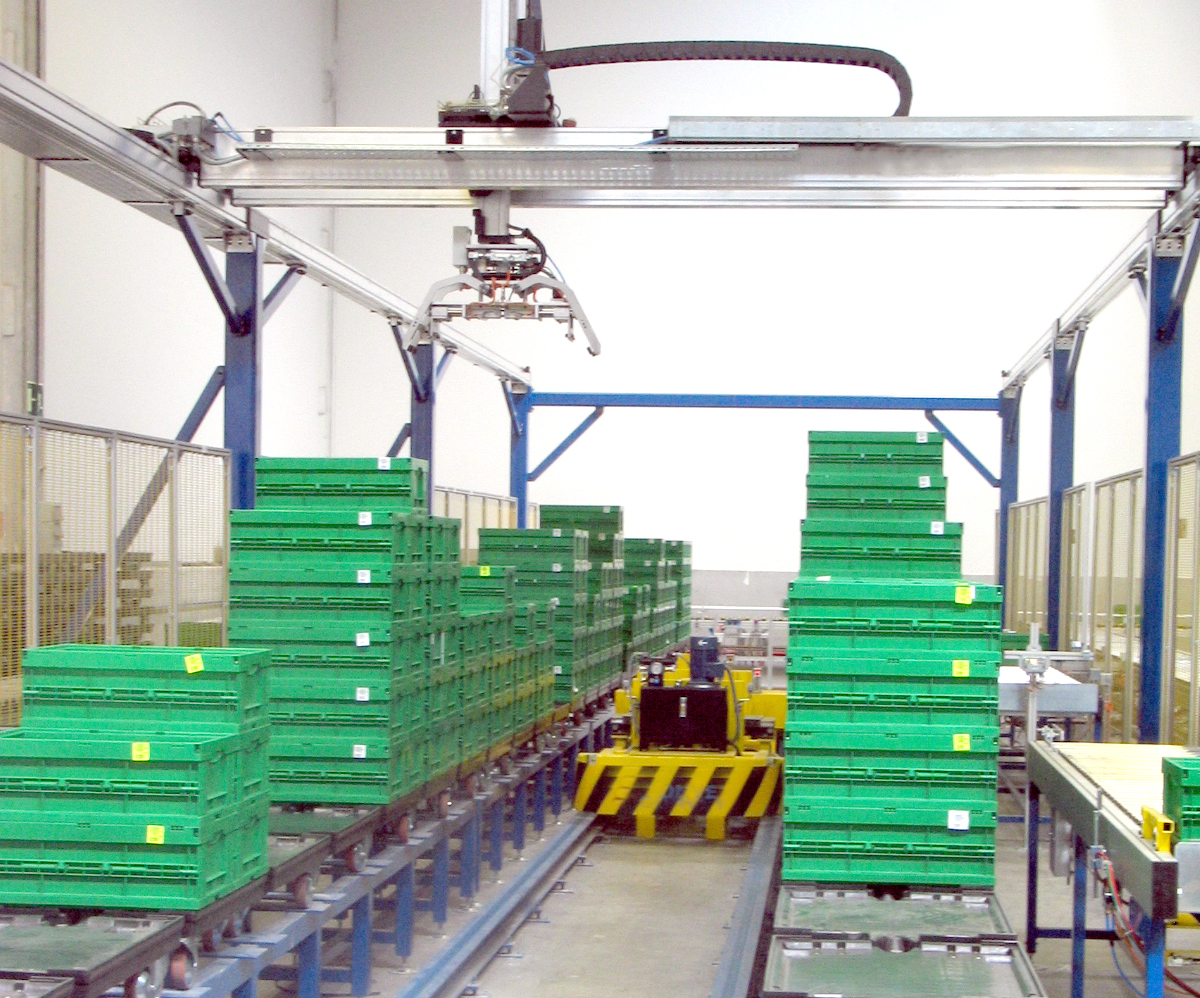}\label{15-2}}
	\subfigure[]{\includegraphics[height=0.16\textheight]{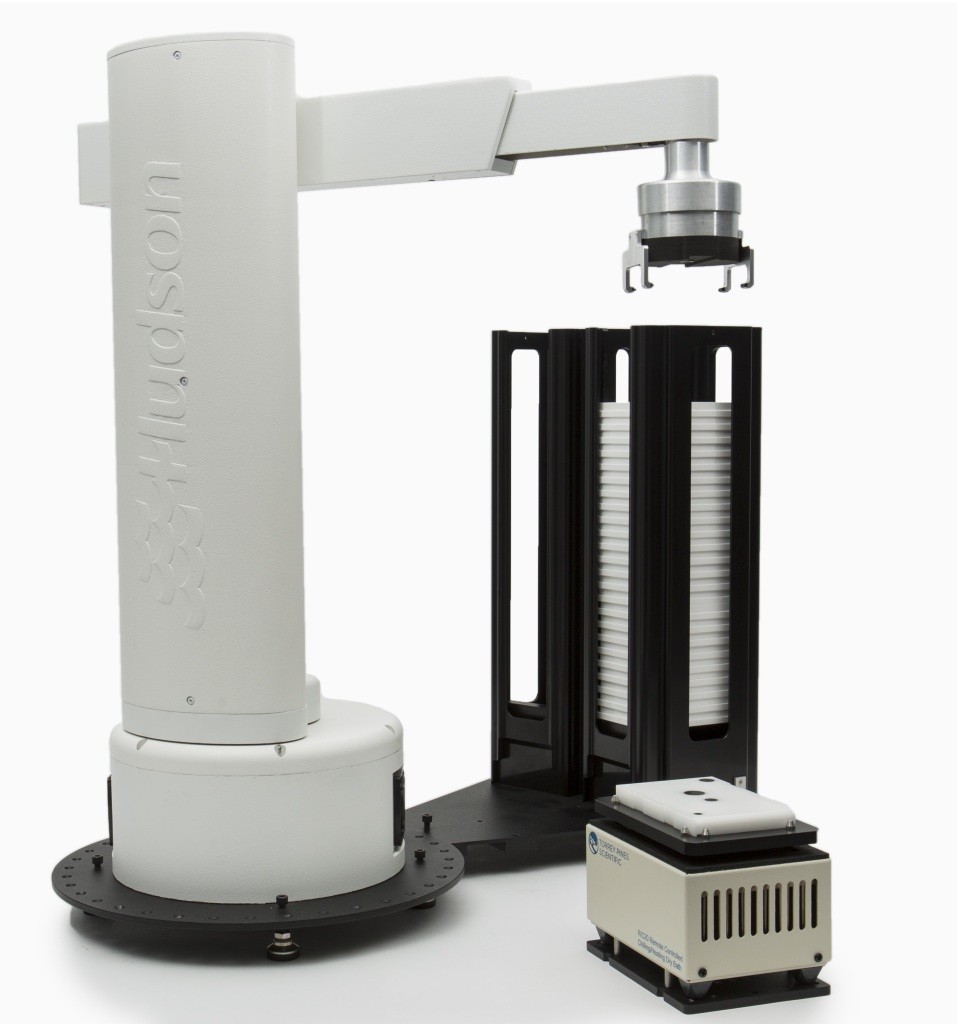}\label{15-3}}
	\subfigure[]{\includegraphics[height=0.16\textheight]{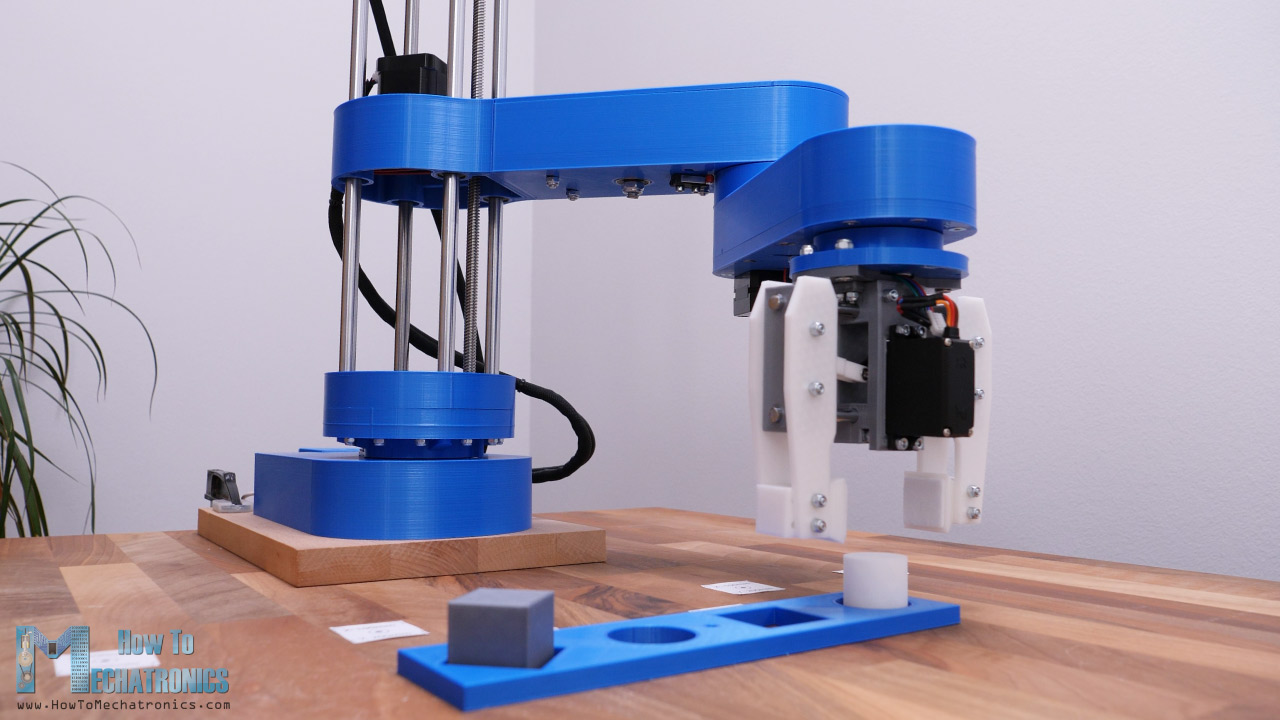}\label{15-5}}
	\subfigure[]{\includegraphics[height=0.16\textheight]{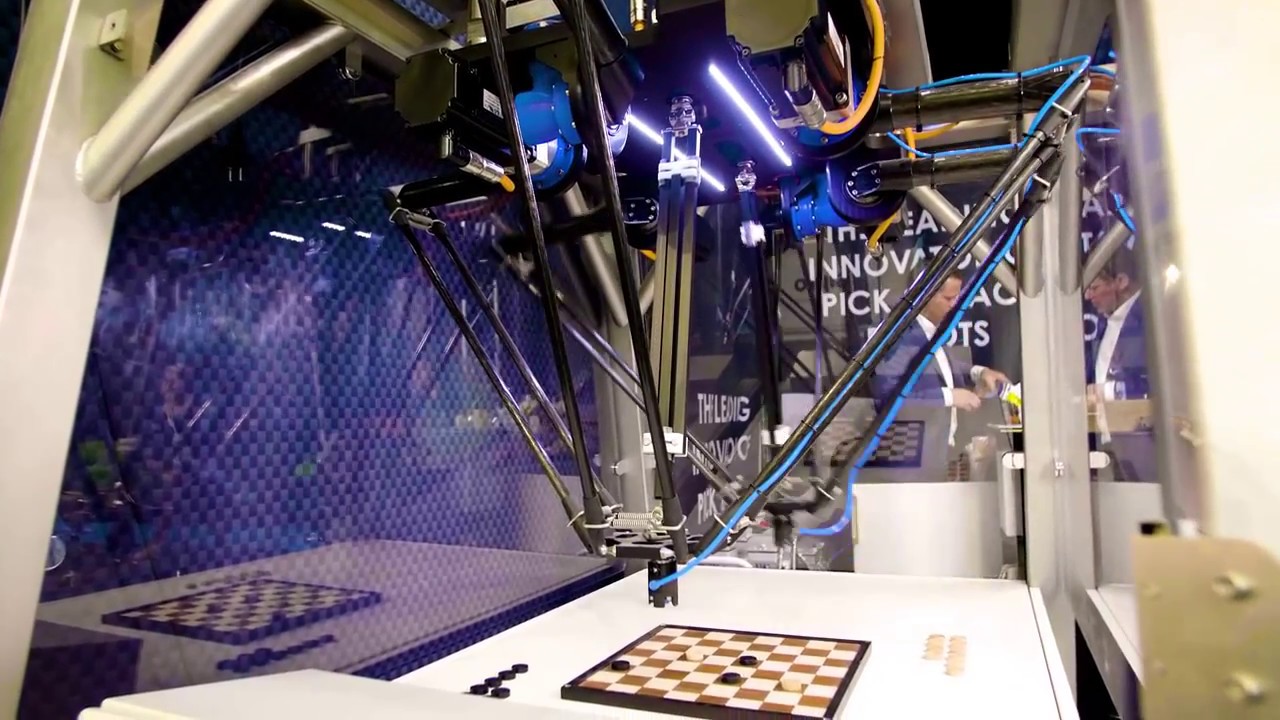}\label{15-6}}
	\subfigure[]{\includegraphics[height=0.16\textheight]{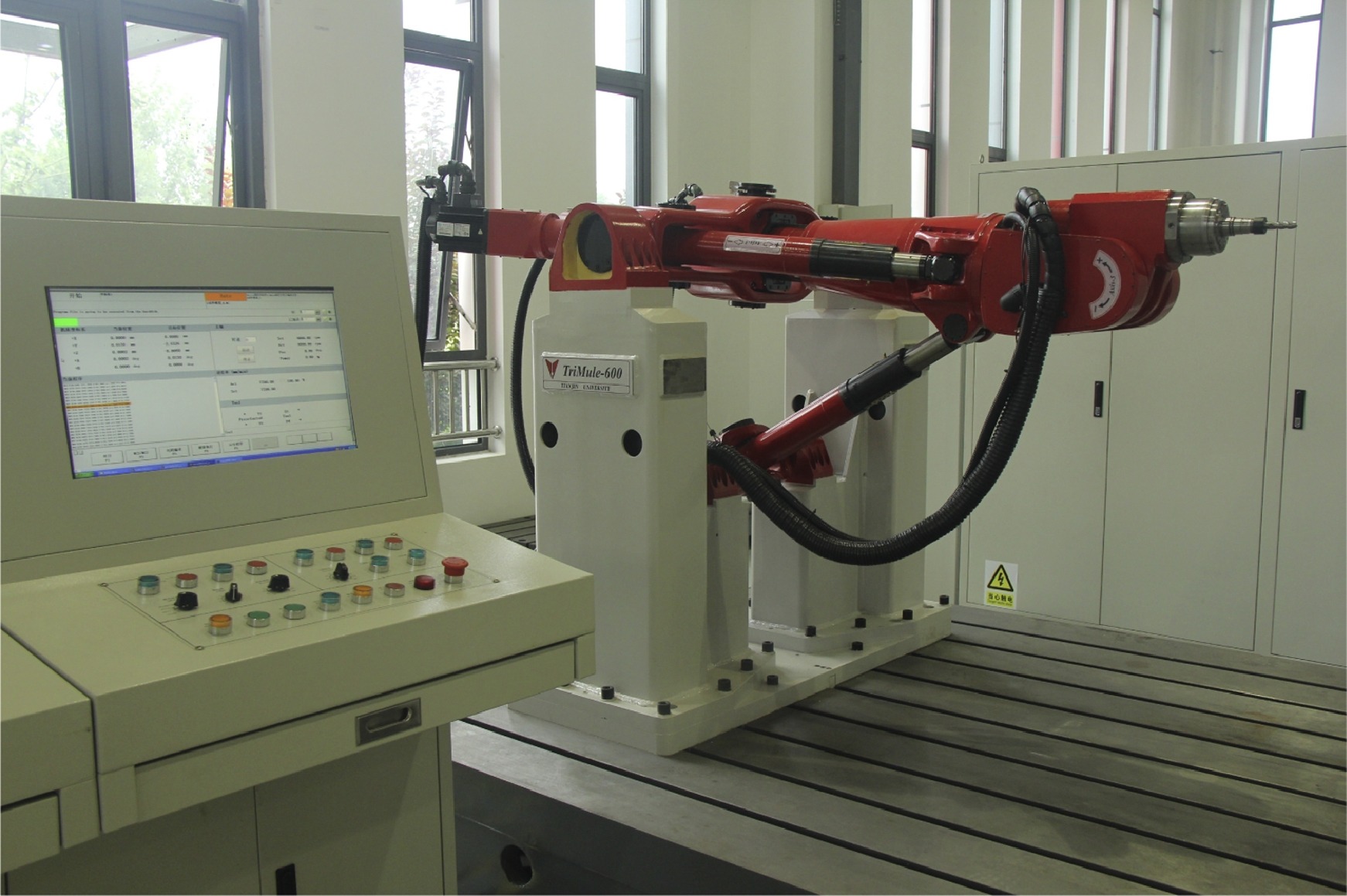}\label{15-7}}
	\subfigure[]{\includegraphics[height=0.16\textheight]{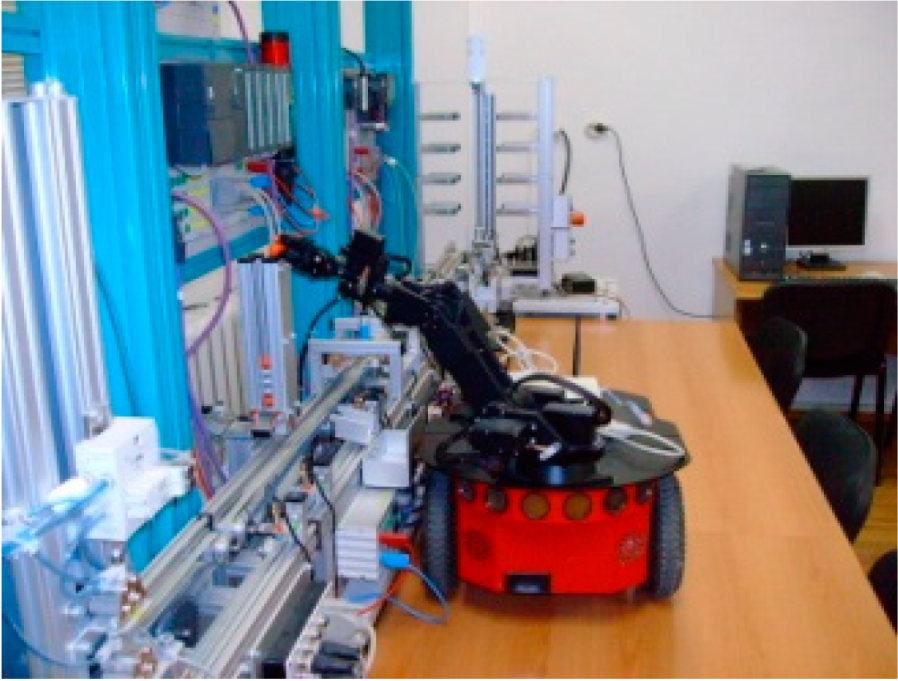}\label{15-8}}
	\caption{Sample industrial robots with applications in \ac{RAD}: (a) A Unimate robot with polar configuration~\cite{spectrum_2018}, (b) An articulated robot in a foundry~\cite{industrial_robot}, (c) A gantry robot~\cite{cartesian_coordinate_robot}, (d) A benchtop cylindrical microplate handler robot for lab automation~\cite{hudson}, (e) A Selective Compliance Assembly Robot Arm~\cite{dejan_nicholas}, (f) A Codian Robotics delta robot~\cite{youtube}, (g) A prototype machine of the TriMule\textsuperscript{\textregistered} robot~\cite{liu2019inverse}, (h) A \emph{Wheeled Mobile Robot} within an Assembly/Dissasembly Mechatronics Line~\cite{dragomir2019modelling}.}
	\label{15}
\end{figure*}

In this section, we will introduce some of the most common types of industrial robots with applications in \ac{RAD}. An overview is shown in Fig.~\ref{15}.\\
\textcolor{blue}{A study of the role of dual-arm robots in manipulation tasks involved in \ac[4]{RA} procedures has been reported in~\cite{weng2019survey}, where the integration of online identification and hardware mechanism, which is required for compliance, as well as perception by means of employing sensors, has been discussed as essential components of such a system.\\}

\emph{Unimate}, the first industrial robot,  was based on a polar, i.e. spherical configuration that, as shown in Fig.~\ref{15-4}, was tailored to the hydraulic drive utilized for powering it. It involved five joints, including two rotary joints at the base and the shoulder, a prismatic joint for extending or retracting the arm, and two more rotary joints at the wrist. Nevertheless, these were not enough for orientating the tool as desired, given that six degrees of freedom were required which was not feasible, due to the incompetence of the relevant control technologies of the time~\cite{wilson2014implementation}.\\
The most prevalent category of today's industrial robots are \emph{articulated}~\cite{diy_2021}. These are often referred to as manipulator arms or robotic arms~\cite{iso}. As shown in Fig.~\ref{15-1}, they are inspired by human arms, and possess the capability of performing various motions, owing to their high degrees of freedom.\\
\emph{x-y-z robots}, which are also referred as \emph{rectilinear, gantry or Cartesian robots}~\cite{osha,diy_2021}, are used in a wide range of applications. They take advantage of three prismatic joints and three rotary joints for achieving the desired spatial position and orientation, respectively (see Fig.~\ref{15-2}). For tasks being carried out in a planar workspace, only three of the total six degrees of freedom suffice, two for the position and one for the orientation~\cite{industrial_robot}.\\
Another category of industrial robots consists of \emph{cylindrical coordinate} ones. They involve a rotary actuator at the base, as well as one or more prismatic joints connecting their links to each other. As illustrated in Fig.~\ref{15-3}, by the virtue of the foregoing structure, they are capable of moving both horizontally and vertically through sliding. Moreover, taking advantage of compact End-Effector designs, they can reach narrow workspaces conveniently~\cite{diy_2021,hudson}.\\
As a class of exemplary \ac{RAD} machines, Selective Compliance Assembly Robot Arm (SCARA) constitute another configuration of industrial robots, which, as shown in Fig.~\ref{15-5}, possess two rotary joints that are parallel to each other, aiming at high-precision motion and positioning within a horizontal plane. They support a rotary shaft which holds the End-Effector in the vertical orientation~\cite{osha}.\\
All categories of industrial robots discussed are serial ones, which notoriously suffer from the accumulation of the errors throughout the chains, i.e. the series of links and joints connecting their base to the End-Effector, resulting in less rigidity with an increased length of the chain. Therefore, parallel configurations also exist, which involve more than one chain.\\
In addition to the fact that the chains involved in a parallel manipulator are relatively simple and short, which makes them less prone to inaccuracies, the fact that their motions, as well as possible off-axis tolerances, are restricted by the rest of the chains in the underlying closed-loop fashion further contributes to preventing them from undesired movements, where the errors are averaged, rather then being accumulated.\\
For example, as shown in Fig.~\ref{15-6}, delta robots posses a parallelogram linkage structure, involving a four-bar configuration, and are suitable for high maneuvering tasks, e.g. fast Pick-and-Place operations, and direct control procedures~\cite{industrial_robot}.\\
The \emph{TriMule}\textsuperscript{\textregistered} is another robot that has been successfully utilized for performing machining, e.g. drilling, tasks on large parts in the context of \ac[4]{RA}, due to its improved stiffness and accuracy~\cite{li2019deformation}. It possesses a compact 5-degrees-of-freedom structure, consisting of a 3-degrees-of-freedom parallel positioning manipulator and a 2-degrees-of-freedom wrist~\cite{yang2018continuous} (see Fig.~\ref{15-7}).\\
In a \ac[4]{RAD} Mecatronics Line, \emph{Wheeled Mobile Robot}, as shown in Fig.~\ref{15-8}, may work in two different ways. One strategy, which is suitable for products requiring transporting parts with various weights, concerns the Wheeled Mobile Robot working in parallel, where e.g. one performs transportation tasks, and the other one carries out manipulations. Alternatively, the robot could be involved in doing both transportation and manipulation tasks, each following its own trajectory, where one robot is responsible for even workstations, and the other one for the odd ones. By doing so, it will be possible to optimize the whole production cycle uniformly. It should be noted that for accurate localization of the Wheeled Mobile Robot, utilizing synchronization signals could be considered, obviating the necessity of visual serving systems~\cite{dragomir2019modelling}.

\emph{Dual-arm robots} have attracted serious attention from industrial and academic communities during the last few decades, and are widely thought to play a significant role in various \ac[4]{RAD} tasks. Bi-manual Baxter robots could serve as a reliable example~\cite{rizwan2019formal}. They are used in multi-robotic construction scenarios as well~\cite{ahmad2019hybrid}.\\
\emph{Cable-driven Parallel Robots}, on the other hand, are greatly flexible, and their modularity and reconfigurability paves the way for their substantial applications in \ac[4]{RAD}. More clearly, the possibility of manually attaching or detaching a desired number of modular branches enables to adjust their spatial topology as required, where modifying the connection points on the End-Effector allows for applicability to a further diversified spectrum of \ac[4]{RAD}-related operations. The vector closed rule and the Lagrange method constitute essential elements in analyzing, understanding and solving the underlying Inverse Kinematic Problem\footnote{Calculation of the required values of the joint parameters for the \ac{EE} to appear at a certain pose with respect to the base~\cite{peiper1968kinematics,ikp}.} using Inverse Dynamic Programming\footnote{The maximum~(minimum) of a problem is the inverse function to the minimum~(maximum) of the inverse problem~\cite{iwamoto1977inverse}.}, which is a prerequisite for making use of the robot in an effective and efficient way~\cite{zhao2019typical}.\\
The usage of the ABB~IRB~140 robot for executing \ac[4]{RA} tasks through a desired trajectory has been reported in~\cite{duque2019trajectory}. Sawyer, which is a robot from Rethink Robotics is a suitable choice when it comes to collaborative tasks.\\
A typical \ac[4]{RAD} setup for Waste Electrical and Electronic Equipment may consist of a two-axis manipulator and a wave soldering machine, which are coordinated for motion using a Central Control Unit, accompanied by a suction cup for picking components~\cite{marconi2019feasibility}.

\subsection{Object Grasping and Manipulation}
\label{sec:execution:object_grasping_and_manipulation}

During \ac[4]{RAD}, the robot has to be able to grasp the part at hand, as well as relocate and reorient it, depending on the nature of the underlying task. In either case, it necessitates keeping track of the part's 6D pose. The most common approach to this problem is vision-based. Setting it up demands making a database of pose examples for traininng a statistical model (most commonly a deep neural network), as well as an evaluation metric describing its accuracy, which should be defined well enough to represent numerous operational and quality-related criteria, including geometric alignment. Expectations need to be adjusted, considering the type of the materials involved, along with the properties of the robot and the gripper. The evaluation metric standing for the reliability of a certain pose estimation instance could be defined based on the conditional probability of successfully completing the task given the current pose~\cite{hietanen2019benchmarking}.

\subsection{Compliance}
\label{sec:execution:compliance}

\begin{figure}[t]
	\centering
	\includegraphics[width=.48\textwidth]{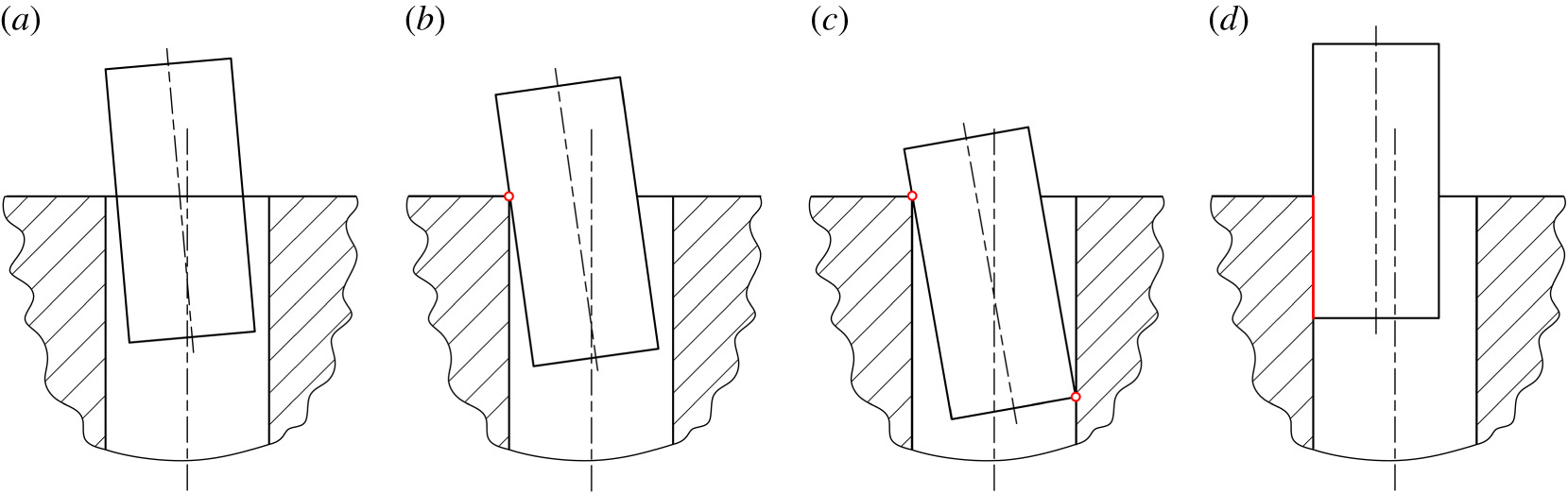}
	\caption{Schematic illustration of a peg being withdrawn from a hole in a disassembly process, involving (a)~Zero, (b)~One, (c)~Two or (d)~Infinite contacting points between them, the latter constituting a line. The figure has been taken from~\cite{zhang2019peg}.
	}
	\label{14}
\end{figure}

Tools and machines with high positioning accuracy are relatively expensive, which rises associated costs, especially in the case of \ac{RA}. Therefore, \emph{compliance}, i.e. \emph{the relationships between the motions and forces produced by the robotic manipulator and the part at their contacting point}, is of paramount importance.\\
Compliance plays a crucial role in applications where the execution of the assembly task requires the presence of certain motion or force profiles, and in contexts where the positioning requirements result in relatively small error tolerances due to e.g. the uncertainties involved.\\
Compliance may also greatly contribute to avoiding jamming and wedging, as well as to achieving robustness against external motions and forces affecting the robotic \ac{EE}\footnote{The device attached to the end of a robotic arm, which is supposed to perform the main task the robot is aimed at, and interact with the work environment~\cite{monkman2007robot,wikipedia_2021}.}, and performing delicate tasks such as window-wiping, which demand both position and force control.\\
Compliance may be modelled using stiffness or compliance matrices, and accomplished through the robotic joints' servo-actuators based on their stiffness, which may be controlled indirectly by manipulating the joint torques. Doing so for achieving feedback control of the robotic \ac{EE}'s force or position, or both, constitutes \emph{active compliance}.\\
The limited stiffness of the parts being assembled and numerous elements of the robotic manipulator, including the gripper, the drive system and the links, may further contribute to the compliant motion, which is referred to as robot \emph{passive compliance}.\\
Both active and passive compliance have advantages and disadvantages, which need to be taken into account, considering the particularities of the certain \ac{RAD} task at hand, in order to devise and implement a suitable compliance strategy~\cite{wang1998passive}.\\
For example, as shown in Fig.~\ref{14}, from the point of view of a 2D cross-section, a cylindrical peg that is being withdrawn from a clearance-fit hole may be in four possible statuses in terms of the corresponding contacts: no contact, single-point, two-point or line.
A two-point contact may lead to wedging or jamming. This is an instance of the cases where active-compliance control could play a significant role. The degree, location and initial error related to the compliance considerably affect the resulting performance. A quasi-static analysis of the foregoing factors, as well as the range, position and boundary conditions of the two-point contact region has been reported in~\cite{zhang2019peg}.

\subsection{Tackling Uncertainties}
\label{sec:execution:tackling_uncertainties}
When automation is utilized as the main strategy for improving the repeatability of \ac{RAD} processes, inherent uncertainties and errors can undermine performance~\cite{jenett2019relative}.
Robotic disassembly may be complicated by various types of uncertainties, e.g. components interlocking, which requires making use of re-planning strategies enabling to detect separable components and subassemblies. The foregoing aim could be achieved in a minimal number of steps through checking for interferences by means of a summation operator and a list with two pointers, followed by optimization for coming up with new subassembly directions and sequences~\cite{laili2019robotic}.\\
On the other hand, one could think of utilizing interval programming, as well as fuzzy or stochastic methods, for tackling uncertainties~\cite{ming2019multi}.\\
A considerable part of the relevant literature has been dedicated to automatically detecting, or even predicting, and then correcting possible errors while performing \ac[4]{RA}. Functional Principal Component Analysis could be used to find out the multi-dimensional feature values associated with force and torque, followed by clustering the results, by means of e.g. the k-means algorithm and decision trees, into either successful or one of the several possible cases of erroneous operation~\cite{hayami2019multi}.\\
In the presence of uncertainties, achieving smooth performance in terms of the tasks allocated to each robot for Multi-Robot stations, modeling a stochastic partial Assembly/Disassembly Line Balancing Problem and solving it as through Multi-Objective Optimization could be considered~\cite{wang2019partial}.\\
On the other hand, various \ac[4]{RAD} tasks may be prone to mechanical errors and inaccuracies, which need to be simulated, accounted for and compensated. For example, orbital drilling processes referred to as helical milling, which are used in aircraft assembly, are susceptible to undesirable performance due to excessive cutting forces and temperatures. Depending on the types of uncertainties involved, different offline compensation measures may be taken. For example, imprecise hole diameters may be alleviated through applying variable eccentricities at different depths throughout the hole~\cite{li2019deformation}.\\
Auto-calibration for improved reliability might be necessary in cases where the mechanical design entails changes of physical variables throughout operations. It may be realized through monitoring the performance by means of one or more cameras or other types of sensors. For example, kinematic uncertainties involved in the performance of redundant Cable-drive Parallel Robots, e.g. the ratio between the actuator angle and the cable length, could be alleviated through using the measurements returned by an encoder, based on a model describing the relationships between the measurement errors and structural parameters~\cite{zhang2019auto}.

\subsection{Simulating Execution}
\label{sec:execution:simulating_execution}

Simulation of \ac[4]{RAD} execution can improve predicting and preparing for uncertain conditions and efficiency~\cite{axelsson2019using}.\\
Various pieces of software have been developed for performing robotic simulations, which is vital for verifying the feasibility of planned \ac[4]{AS}. One example is the Box2D physics simulator~\cite{geft2019robust}. Blender and SolidWorks are instances of relevant software which could be used for 3D modeling. Virtual assembly and scene reconstruction could be performed using EON Studio~\cite{wen2019research}. This category of software utilized for designing architectures is referred to as Computer-Aided Architectural Design. Teamcenter and Vis MockUp are other pieces of software that could be used for 3D visual simulation in the context of robotic assembly~\cite{hui2019research}.\\
Gazebo provides a simulation environment which could be utilized for training for \ac[4]{ASP}~\cite{zhao2019aspw}. RobotStudio could be used for simulating \ac[4]{RA} operations~\cite{duque2019trajectory}.\\
SAMCEF is suitable for Finite Element Analysis\footnote{Analyzing problems using the finite element method, which solves differential equations numerically in two or three state or boundary variables~\cite{logan2012first,fea}.} of assembly tasks such as drilling, in order to predict and account for possible uncertainties~\cite{li2019deformation}.\\
The Industrial Path Solutions software~\cite{hanson2019industrial}
is a suitable choice for developing virtual models of operations environments, especially when it comes to analyzing the application of \ac[4]{HRC}~\cite{axelsson2019using}.

\subsection{Human-Robot Collaboration}
\label{sec:execution:hrc}

Although \ac[4]{RAD} may be extremely efficient in handling repetitive, routine, strategic or general tasks with prescribed forces and relatively high spatial precision and power, it may not be flexible enough to handle changing task requirements, complicated jobs, non-serialized products or large parts. These are requirements for fast and low-cost production, and should be made symbiotic, immersive, dynamic, fast and safe to be attractive for industrial parties, who could utilize such systems with strong control interfaces~\cite{pecora2019systemic,perez2019symbiotic}.\\
This motivates \ac[4]{HRC} for \ac[4]{RAD}, which requires certain considerations in terms of \ac[4]{SP} for the satisfaction of requirements such as a suitable workload balance~\cite{xu2020disassembly}. In fact, the main advantages of a human operator collaborating with a robot in the context of assembly or disassembly arise from the intelligence and dexterity brought by the human, which combined with the aforementioned virtues of robots, make \ac[4]{HRC} systems greatly flexible, accurate and reliable. Nevertheless, allocating and balancing the tasks based on the available skills and a classification of the tasks based on the level of complexity~\cite{malik2019complexity} for achieving the best possible performance gives rise to an additional Process Planning challenge~\cite{dianatfar2019task}.\\
In the presence of various types of uncertainties, e.g. in the context of disassembling press-fitted products, human-robot collaboration takes essential importance, and could be realized through active compliance control~\cite{huang2019case}. It is substantially useful for improving the flexibility, efficiency and applicability of robotic disassembly, as well as alleviating the idle time and the manual burden on the human collaborating with the robot(s), due to the further intelligence and assistance brought by human-robot collaboration through the knowledge base made using Natural Language Processing\footnote{Programming computers to understand concepts, as well as the nuances associated with the context, from large volumes of documents representing natural human language, as well as organizing and categorizing them~\cite{kongthon2009implementing,nlp}.} from disassembly relationships, provided as a robotic knowledge graph, which is helpful for obtaining, analyzing and managing the disassembly data~\cite{ding2019robotic}.\\
It should be noted that one of the factors that could affect the accuracy of \ac[4]{HRC} for \ac[4]{RAD} tasks is human fatigue, which more frequently occurs during processes of disassembling products of the same type, and needs to be duly modeled and taken care of~\cite{li2019sequence}. Moreover, the workers' level of experience and proficiency, as well as structural similarities between different products, may have an impact on the operational requirements and limitations~\cite{xia2019balancing}.\\
Brain-Computer Interfaces may also be utilized for enhancing the communication between the human and the robot. It leads to higher flexibility in handling situations where the workspace of the robot has been largely affected by external constraints or where the operator needs extra external assistance from another human operator. This can be achieved by using wavelet decomposition for extracting certain sub-bands of ElectroEncephaloGraphy (EEG) signals which may be characterized based on the underlying frequency, e.g. 0–4~Hz or 4–8~Hz. The Pearson correlation coefficient could be utilized to find out the relation between the 14 pairs of EEG channels. The validity of brain network parameters provide an indication of the motion characteristics. The eye movement characteristics could be analyzed based on the information provided through certain channels.
Using the foregoing strategy, it is possible to detect intended left and right motions with an accuracy of about 97\%, and improve the efficiency by about 60\% compared to the case where another human operator performs the task of assistance, instead of an auxiliary robot~\cite{huang2019study}.\\
Since human operators are always prone to various types of errors caused by e.g. fatigue, their safety constitutes a major concern in the design of \ac[4]{HRC} for \ac[4]{RAD} systems. This could be analyzed based on a formal non-deterministic model. In this way, a safety engineer could detect all possible cases of dangerous, erroneous human actions, and revise the design over multiple iterations, until all of them are detected and accounted for~\cite{askarpour2019formal}.\\
Other measures intending to improve the safety in the context of \ac[4]{HRC} consider monitored regions according to which the motions of the robot could be constrained, as well as utilizing safety skins~\cite{papanastasiou2019towards}.\\
Developing a reconfigurable, continuous co-share environment for \ac[4]{HRC} may benefit from software decision mechanisms, redundant transmission protocols and management frameworks for unifying various resources of sensing~\cite{pecora2019systemic}.\\
On top of safety, for several other reasons, it is vital to equip robots used in \ac[4]{HRC} with cognitive skills utilizing hybrid conditional planning. Mainly, the robot should be capable of handling high-level planning for \ac[4]{ASP} and \ac[4]{DSP}, as well as geometric feasibility checks. Moreover, the robot needs to be capable of performing communication, sensing and commonsense reasoning~\cite{rizwan2019formal}.\\
Operator task allocation is an essential component of the process of Cooperative  Assembly/Disassembly  Sequence  and  Task  Planning, which determine the task sequence of each operator, as well as the time taken to perform each sequence. They could be performed using Genetic Algorithms, based on the associated Disassembly Hybrid Graph Model~\cite{tian2019product}.\\
A major benefit of \ac[4]{HRC} arises from the opportunity to use remote lead-through control strategies based on a Cyber-Physical System\footnote{A system in which a mechanism is monitored or controlled using computer software~\cite{cps}.}, possibly using a model-based display, which obviates the necessity of a human operator being physically present in a hazardous operational site, such as dusty, high-pressure or high-temperature environments~\cite{liu2020remote}. For example, in the context of coating industry, the application of a lag-removing robot for zinc pot on a hot-dip galvanizing line has been reported in~\cite{bai2019application}.\\
\ac[4]{HRC} systems employed in \ac[4]{RAD} applications may take substantial advantage of gesture recognition technologies for a more smooth and natural performance, through which the human operator guides the robot throughout the course of performing the relevant tasks. A gesture recognition framework requiring a single shot has been proposed and tested on the JiaJia social robot in~\cite{kuang2019one}, where Dynamic Time Warping is utilized for temporally aligning two sequences of gestures based on a frame-specific indicator of distance representing spatial features. Moreover, attention regions are extracted in a two-directional manner, which allows to retrieve the necessary data from both static and dynamic, i.e. hold and movement, parts of each gesture.\\
Continuous online recognition of gestures presented by multiple users may lead to improved robustness, compared to cases where a separate database is used for each user. The sensory information could be obtained using non-interfering equipment, e.g. sensors mounted on the tools and depth cameras, which has been realized with precision and recall rates of about 90\% in~\cite{coupete2019multi}. In the foregoing context, the choice of features and focusing on the most effective ones, which are usually concentrated on hand movements, may play a major role in achieving a desirable recognition performance.\\
Force control, on the other hand, plays an essential role in \ac[4]{HRC}-based \ac[4]{RAD} frameworks, where the traction force inserted by the hand and the contact force between the manipulated object and the End-Effector are captured using six-dimensional sensors, e.g. tandem type ones, followed by interpreting the data in order to determine the motion intended as a part of the guidance~\cite{zhang2019six}.\\
Dynamic hybrid position/force controllers are useful for performing \ac[4]{RA} tasks with Multi Degrees of   Freedom manipulators, which may be redundant, modeling the constraints on the End-Effector as sets of hypersurfaces~\cite{wang2019dynamic}.\\
On top of force control, other communication modalities such as vision systems aimed at detecting and tracking objects and air-pressure sensors, as well as the ones provided through wearable devices, including Augmented Reality technologies, haptic interfaces and audio feedback, may contribute to the ease of \ac[4]{HRC} for \ac[4]{RAD}~\cite{papanastasiou2019towards}.

\subsubsection{Learning from Human Demonstrations \label{2}}
It is possible to make a robot learn Pick-and-Place\footnote{Fast, high-precision placement of a wide range of electronic \acp[4]{SMD}, including integrated circuits, resistors and capacitors, as well as through-hole components, onto \acp[4]{PCB} utilized in telecommunication, medical, military, automotive and industrial devices, consumer electronics and computers~\cite{pick_and_place_machine}.} policies to be employed for assembling a product using a demonstration of its disassembly. This may help facilitate the task for non-expert human users who do not find engaging in computer programming convenient. The foregoing procedure requires making use of task-tailored training or 3D  Computer Aided Design models for pose and orientation estimation. Enabling the robot to generalize to other objects could be realized through shape matching processes aimed at learning shape description priors, which visually represents the geometric correspondence between the surface of the object and that of its target location. The foregoing functionality could be achieved in a self-supervised manner, and allows to obtain a more widely applicable perception of the task in terms of how parts of different shapes and associated surfaces fit together in the context of assembly~\cite{zakka2019form2fit}.\\
\begin{figure}
\centering
\includegraphics[width=0.45\textwidth]{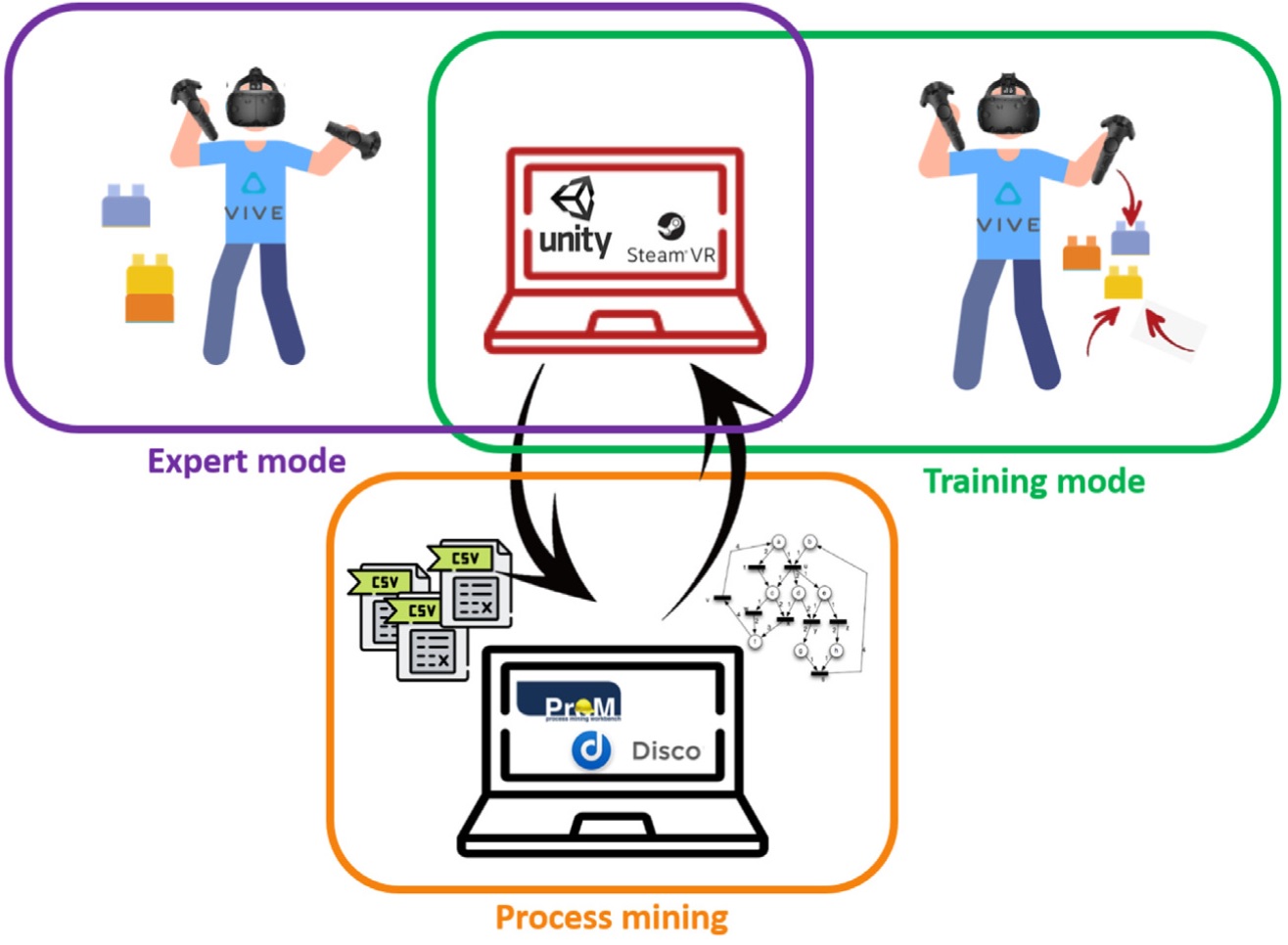}
\caption{System diagram: expert mode allows the expert operators to introduce their knowledge into the system. Process mining automatically generates the assembly models from the operation logs, and training allows the new operators to learn the assembly~\cite{roldan2019training}. }
\label{fig3}
\end{figure}
Learning by demonstration techniques, such as  Task Parametrized Gaussian Mixture Model, contribute to leveraging the complexity and time-consumption involved in programming for the trajectory generation and execution of \ac[4]{RA} operations. The demonstrations could be captured using a motion sensor such as Kinect, and then supplied into the training module using a probabilistic model. Subsequently, Petri Nets\footnote{A family of discrete dynamic systems utilized for mathematical modeling of distributed systems, where places and transitions are shown as white circles and rectangles, respectively, within a bipartite graph~\cite{pn}.} could be utilized to automatically generate assembly plans for the parts detected from the sensory data~\cite{duque2019trajectory}.\\
In order to reduce the number of demonstrations that are required for training the robot to perform a certain task, it is possible to use neural dynamics. In this case, a single demonstration may suffice for collecting the necessary information regarding the underlying sequence of sub-goals and the time intervals between them. The robot could, in turn, instruct non-expert human operators based on what it has learned about the sequence of tasks, with an adjustable time scale~\cite{cunha2019towards}.\\
Finally, Virtual Reality could be utilized to enable expert users to transfer their knowledge and experience to other operators through process mining~\cite{roldan2019training}. This process is illustrated in Fig.~\ref{fig3}. 
\section{Facilitating Planning and Execution by Design}
\label{sec:facilitating}

\label{sec:design_for_assembly}
Assembling a product comprising of $N$ parts typically consists of $N-1$ steps. Reducing the number of parts by design improves efficiency and reduces costs~\cite{murali2019integrated}.\\
Discrete elements may contribute to developing modular ultralight lattice structures, which facilitate the manufacturing and \ac[4]{RA} of elements. Furthemore, they can be possibly further expanded, disassembled or reassembled to larger structures of various types. 
The resulting structure will also be of improved navigability by small mobile robots, as well as enhanced structural efficiency, where the wasted payload volume could be minimized as a result of the upgraded packing efficiency. 
It should be noted that these properties are affected by geometric variables such as the face, edge and vertex connectivity between the unit cells, along with the density of the attachments over a unit of area or a given number of voxels. In the following, we briefly discuss desirable mechanical properties of resulting structures~\cite{ochalek2019geometry}. 
\subsubsection{Stiffness and Strength Scaling}
The power by which stiffness and strength scale with respect to the density, i.e. the fraction of the space being occupied by the solid part of the material~\cite{fleck2004overview}, are of paramount importance in determining the extent to which it may comply with the requirements of an \ac[4]{RAD} framework. Such properties are affected considerably by the cell wall bending of the cellular structure, and may be evaluated using dimensional analysis in conjunction with beam theory. Certain geometric shapes have been studied in terms of the foregoing scaling in the literature, and those that are not may be, at least roughly, evaluated based on the approach introduced in~\cite{deshpande2001foam}, which relates the foregoing properties of lattice structures to the connectivity of the unit cells, depending on its microstructural characteristics being either axial stretch dominated or transverse.
\subsubsection{Assembly Efficiency}
\label{sec_Assembly_Efficiency}
An efficient \ac[4]{RA} system needs to be fast and low-cost. Moreover, the tolerance required seriously affects the resulting efficiency. Furthermore, assuming that a standardized manner of utilizing fasteners exists and has been adopted, the number of fasteners required for performing \ac[4]{RA} is also considered an important factor in assessing the efficiency of the overall system.
\subsubsection{Tiling}
\label{sec_Tiling}
\begin{figure}[t]
	\centering
	\includegraphics[width=.48\textwidth]{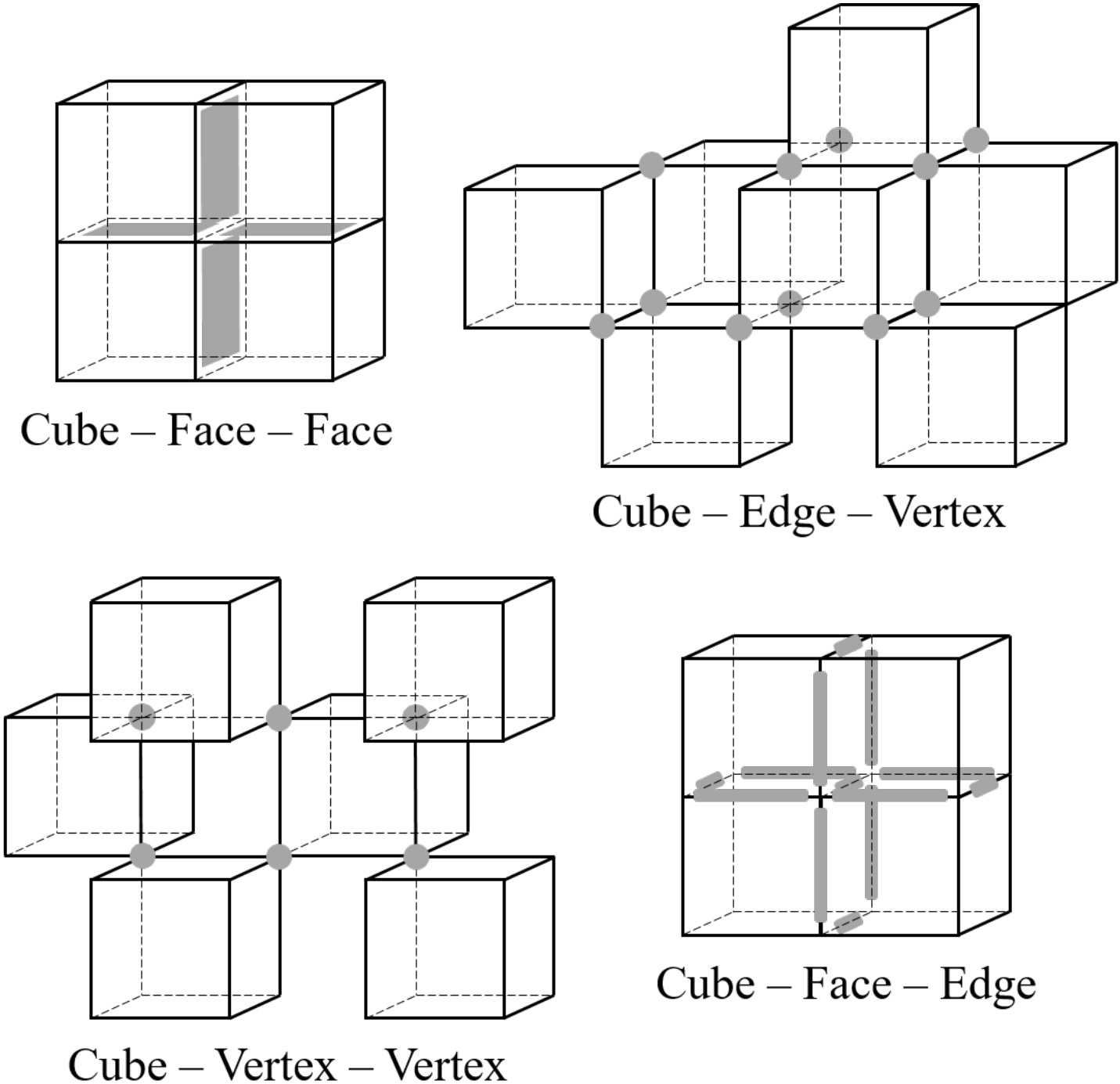}
	\caption{Schematic illustration of various packing and tiling strategies for unit cells that have been fastened to each other. In the above geometric configurations, the attachments re shown in gray, where an adjacency is defined a neighboring voxel's location with respect to the origin voxel, and an attachment stands for a fastener's location. The sub-figures to the top-left and bottom-right represent \emph{Simple Cubic} packing, while the ones to the top-right and bottom-left illustrate \emph{Body Centered Cubic} tiling. The figure has been taken from~\cite{ochalek2019geometry}.}
	\label{3}
\end{figure}

Noticing its effect on the level of complexity of the required robot locomotion to move across the structure, the packing strategy considered for the unit cells that have been fastened to each other, i.e. the spatial relationship between the unit cells, also becomes a major factor in determining the efficiency. As illustrated in Fig.~\ref{3}, options include Simple Cubic packing, which stands for the consecutive layers of the unit cells laying directly on top of each other, and Body Centered Cubix tiling, referring to alternating layers sitting offset to each other.
\subsubsection{Packing Efficiency}
\label{sec_Packing_Efficiency}
The packing efficiency may be formulated as the volume of completed lattice, i.e. deployed volume, to un-assembled voxels' payload volume.
\subsubsection{Volume Allowance for Robotic End-Effector}
\begin{figure}[t]
	\centering
	\includegraphics[width=.48\textwidth]{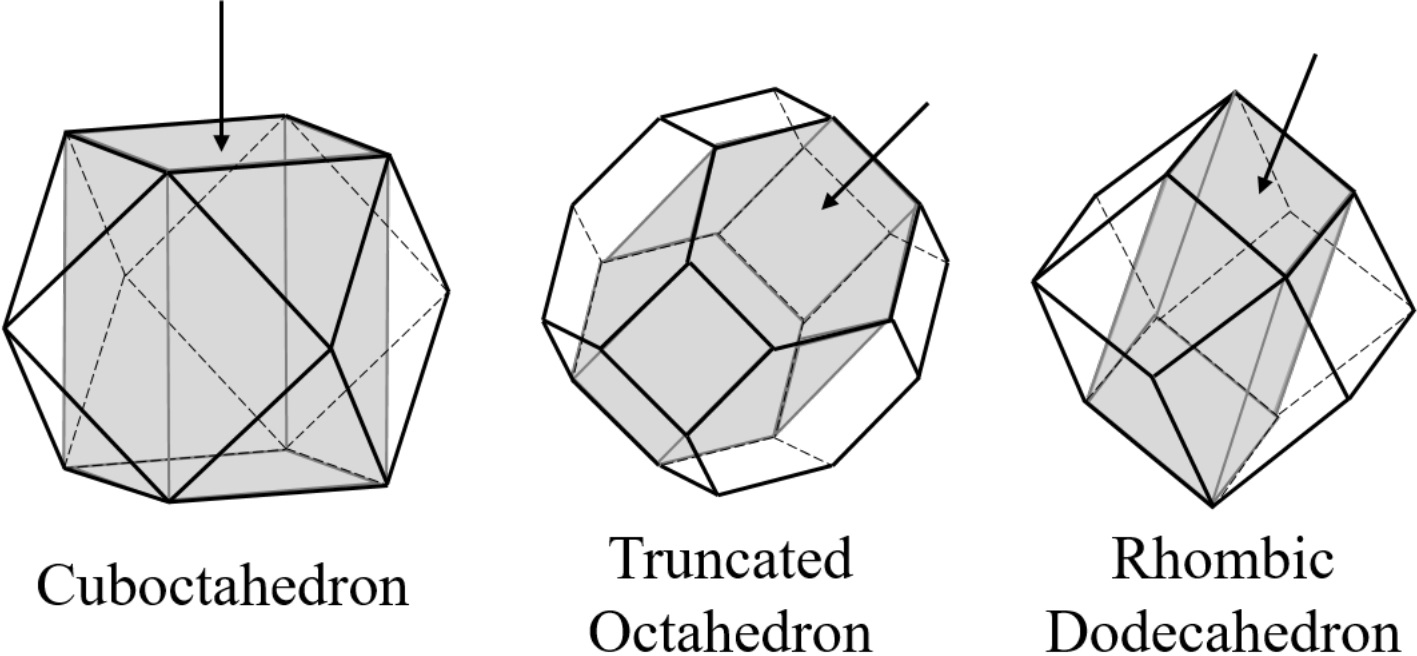}
	\caption{Illustration of the clearance volume, which represents the greatest volume within which the robot's \ac[4]{EE} may enter or exit a cell for assembly. It's ratio to the total volume of the cell is referred to as volume allowance. It is shown in gray, for different unit cells. Arrows show the direction of motion. Figure taken from~\cite{ochalek2019geometry}.}
	\label{4}
\end{figure}

As shown in Fig.~\ref{4}, the ratio of the largest clearance volume through which the robotic \ac[4]{EE} may enter or exit a unit cell to the whole volume of the cell, which is referred to as volume allowance, and determines the level of accessibility of the corresponding attachments, is an essential indicator of the efficiency, in that it is vital for providing the robot with a range of motion being large enough to enable it to maneuver and reach the unit cells and attach them to each other.
\subsubsection{Strut Clearance Angle}
Another parameter affecting the clearance provided for the robotic \ac[4]{EE} to operate near a node is the strut clearance angle, which is defined as the angle between the adjacent strut and the vector orthogonal to the node.

Holding and aligning fixtures is an essential part of every \ac[4]{RA} and Pick-and-Place procedure. Manual design of fixtures may practically take numerous iterations, even for professional designers, which motivates automatic design of cut-outs based on parametrized imprints through dynamic simulation platforms.

\section{Sustainability and Circular Economy  Considerations}
\label{sec:circular_economy}
We have provided a review of characteristics of \ac[4]{RAD} practices in re-manufacturing and recycling processes, as well as a review of their related design and operational challenges and corresponding problems. As \ac{CE} and resource conservation serve as one of the major drivers behind re-manufacturing processes, it would be essential to establish the net contributions, i.e. benefits versus impacts, of \ac[4]{RAD} applications. In this sense, there is a need to assess economic, environmental, and social benefits of \ac[4]{RAD} processes in re-manufacturing and recycling, and integration of these benefits into a sustainability assessment framework, as a means of performance analysis, using multiple criteria analysis~\cite{li2019multi} or integrated performance matrix-based~\cite{kroll1999disassembly} approaches.\\
From an economic perspective, the demand for re-manufactured products and components is increasing. To stay competitive, the re-manufacturing firm has to offer appealing prices which requires stringent measures to reduce cost. Unfortunately, the complexity of disassembly processes in dealing with multiple steps and varying assembly configurations could result in extended disassembly time as well as increased costs~\cite{kroll1999disassembly}. As such, there is a need to industry benchmarks as to what level of complexity versus cost corresponds to a break-even point for adoption of \ac[4]{RD} applications, in which an economic gain, i.e. profitability, is still foreseeable for \ac[4]{RD}-based re-manufacturing~\cite{kimble2020benchmarking}. This margin of profit could be leveraged to justify a partial transfer of the responsibility to manufacturers in accepting additional costs for modifications in design and production process to facilitate disassembly~\cite{reap2002design}. In the long term, the performance and quality of \ac[4]{RD} practices could also be subject to compromises due to degradation in reliability, as well as the precision of robots, and resulting disassembly errors and disruptions, requiring a service life asset management approach for robots and their components~\cite{murray2011analysing,melnik2017improving}. Further, the economic viability of \ac[4]{RD} applications relies on their economy-of-scale, i.e. scale-ability, and their robustness and applicability to a diverse pool of used products and assemblies~\cite{das2000approach}. In this regard, there is a need to develop Key Performance Indicators to trace and analyze the economic efficiency of RD practices over-time subject to changes in configurations, volume, and quality of assemblies. Such indicators could include the \textit{ratio of manual to automated disassembly time}, \textit{ratio of operations requiring tool-change}, \textit{unit disassembly time}, \textit{marginal cost of disassembly}, and \textit{disassembly efforts index}~\cite{das2000approach,li2019multi}.\\
With respect to environmental benefits, RD practices face challenges in terms of maintaining and improving their performance in reuse of disassembled components, i.e. re-manufacturing, recovery of materials, i.e. recycling, and proper segregation of hazardous materials for disposal, due to increasing quantity and diversity of used products, and their varying end-of-life qualities~\cite{renteria2019human,li2019multi}. The carbon emission inventory of \ac[4]{RAD} practices shall be a part of the performance assessment measures, as the collection, transport, sorting, and processing of used products results in carbon emission. More precisely, \ac[4]{RD} processes could be energy intensive due to direct use of energy, e.g. melting and recovery of metals from electronics boards, or extensive repetition of mechanical operations~\cite{guo2020multiresource}. In this regard, \textit{mass recovery ratio}, \textit{emission rate per unit of disassembly}, and \textit{energy intensity per unit recovery} have been suggested as examples of environmental performance assessment of \ac[4]{RD} practices~\cite{li2019multi,renteria2019human}. In this regard, the environmental performance can only be accurately assessed by adopting a life cycle analysis approach considering the direct and embodied emission inventories resulted from manufacturing of robots and their parts alongside the inventories or savings in \ac[4]{RD} processes~\cite{moyer1997environmental,kuo2006enhancing}.\\
\ac[4]{RAD} practices for re-manufacturing are a host of several social and societal challenges. There are various approaches in measurement and assessment of social performance of production processes, including Stakeholders’ theory, institutional theory and resource-based view, socio-technical transition theory, among others~\cite{padilla2020addressing}. Quality control and compliance with customers’, and workers’, health and safety requires adoption of specific procedures and benchmarks given the lack of industry wide standards~\cite{arredondo2019impact}. There exist a trade-off between product customization and capital, as well as operational cost of \ac[4]{RAD}. Flexibility in \ac[4]{RAD} processes for re-manufacturing is an essential attribute to promote users’ appreciation and generate steady demand. Provided the capital intensity of robotic assembly and disassembly lines and setups, many of such processes are task-based and have limited robotic capacities to support customization and diversification of products~\cite{das2000approach}. In case of literature on human-robot interaction, the focus has been on human factors considerations in design and operation of robotic processes. In case of \ac[4]{RAD} for re-manufacturing, the working environment is more prone to exposing workers to toxic and hazardous materials arising from recycling~\cite{gerbers2018safe,dalle2020end} and poorer workplace air quality due to released particulate matters during \ac[4]{RD} process, as well as risks of accidents and injuries due to (unexpected) robot motions~\cite{robla2017working}. These conditions could affect the performance of workers and hinder the quality of their interactions with robots~\cite{arredondo2019impact}. In the long term, the main societal benefits of \ac[4]{RAD} include increased landfill diversion, resource, i.e. material, conservation and efficiency by promoting automation, precision, and standardization of re-manufacturing processes, as well as creation of local jobs related to collection, sorting, and monitoring of re-manufacturing process. As such, the \ac[4]{RAD} practices are in need of developing indicators to assess and report on their performance in line with social dimensions. These indicators could be \textit{risk-based}, such as the likelihood of workers’ exposure to hazardous substances per person per re-manufactured product~\cite{gerbers2018safe,dalle2020end}. There are also \textit{impact-based} indicators presenting the social and societal benefits of the \ac[4]{RAD}-based re-manufacturing, such as the amount of landfill diversion per \ac[4]{RAD} process per year~\cite{renteria2019human}.

\section{Conclusions and Future Work}
\label{sec:conclusion}
Integrating the whole \ac[4]{RAD} pipeline into an end-to-end system such as click-and-assemble~\ac[4]{RA}~\cite{fakhurldeen2019cara} is a promising direction. It should enable the user to operate the system remotely using solely a Computer Aided Design file, where the necessary calculations and preparations for e.g. \ac[4]{ASP} are performed automatically.\\
The fourth industrial revolution has benefited from various aspects of traditional and industrial manufacturing platforms, machinery functionalities and practices, along with the advantages brought by recent smart technologies, altogether forming what is referred to as \emph{smart manufacturing}.\\
Ideally, such a \emph{smart} system should be capable of self-diagnosing the problems, and perform self-monitoring, thereby obviating the necessity of the presence of a human operator, and enhance the level of automation through more efficient data exchange strategies made possible by means of \ac[4]{AI}, cognitive computing, cloud computing, industrial Internet-of-Things and Cyber-Physical Systems (CPS). Internet-of-Things\footnote{Enabling computers to obtain useful data from, and about, the environment, i.e. `things', without requiring human aid, thereby making it possible to acquire information more accurately, rigorously and extensively~\cite{ashton2009internet}.} for manufacturing, Internet-of-Everything\footnote{Further involving other elements, such as people, data and processes, resulting in a more comprehensive concept than the Internet-of-Things~\cite{i_2020}.}, Manufacturing 4.0, smart factory, connected enterprise and industrial internet often indicate a similar, if not the same, concept.\\
A smart factory resembling Industry~4.0 employs a modular structure, where a CPS monitors physical processes, and decisions are made in a decentralized manner. In the foregoing context, as participants of a value chain, CPSs communicate and collaborate with each other, as well as with human beings, both within ad across organizations, over the Internet-of-Things.\\
There are a variety of notions which could potentially open further windows to more highly reliable platforms in the context of Industry~4.0. For example, such frameworks could be equipped with a capability of performing predictive maintenance, which would prevent anticipated losses or damages due to e.g. machinery failures. On the other hand, if realized, since the system utilizes the Internet-of-Things, appropriate use and processing of the data returned by the relevant sensors could enable various organizations participating in the value chain help each other to detect and alleviate their problems in a more cost-effective way through e.g. predicting and notifying other participants of their expected faults and failures.\\
One of the technologies being supposed to play a major role in the advancement of Industry~4.0 is \emph{3D printing}, which is widely believed to facilitate the design process, due to its versatility and flexibility in printing a wide range of geometric structures. Moreover, it is environmentally fairly friendly, and especially for smaller production platforms reduces the lead time, as well as the average production cost. Furthermore, it decreases the warehousing expenses, and helps enable mass customization. Last but not least, 3D printing could make it possible to print spare parts locally, which may exempt a user from ordering and waiting for the parts to be delivered and installed by the provider~\cite{industry4,smart_manufacturing}.\\
Assembling complex and customized products incentivizes the development of systems involving actors of various types, which need to cooperate with each other through e.g. an intelligent CPS-based could framework. It is worth noticing that with increasing contributions of AI to \ac[4]{RAD}, it is expected that sensory information acquired from the real production environment will enrich the assembly or disassembly knowledge model, helping search the available instance library to detect a similar product structure, and guide the process based on the prescribed priority rules~\cite{peng2019knowledge}.\\
Furthermore, one possible direction for future research could involve trying to develop strategies for estimating the required time for \ac[4]{RAD} of given products automatically~\cite{hu2019charts}.\\
The self-reconfiguring cleaning robots discussed throughout the paper could be extended to various other applications as well. For example, the robot aimed at cleaning solar panels could be extended and used for floating solar panels, solar rooftops and solar farms~\cite{chailoet2019assembly}.\\
Regarding Knowledge Transfer, representations of assemblies involved in the present topology might not generalize to actual modular products or LEGO block-based 3D structures involving different practical requirements arising from e.g. non-perpendicular angles between the profiles or nonparallel surfaces. Thus the parts, as well as the relevant restriction, need to be interpreted in terms of surfaces, along with the relationships between them~\cite{rodriguez2020pattern}.\\
In fact, some of the most important advantages of \ac{RAD} in Industry~4.0 arise from their contributions to \ac{CE}, which is essential for climate-change considerations and sustainable development. Further development over various aspects of automatic or semi-automatic \ac{CE}-targeted \ac{RAD} solutions, including \ac{HRC} and Assembly/Disassembly \ac{SP}, is left upon future research for both academic and industrial parties.\\
Nevertheless, not all countries engaged in robotic-related research have yet started paying attention to the technological development of \ac{RAD}. \ac{RD} requires even more serious attention. Despite the prevalent conception, in reality, simply reversing the associated \ac{RA} process may not be feasible due to e.g. geometric constraints or joint specifications, which makes the importance of Design for Disassembly stand out, among other considerations~\cite{poschmann2020disassembly}.\\
One of the concepts requiring further attention in future studies is a systematic review of the existing robotic technologies, as regards their applicability to, and suitability for, specific \ac{RAD} applications. More clearly, the efficiency of a \ac{RAD} process depends, among other factors, on the effectiveness of the robot in performing the tasks planned, which, hypothetically, can be taken into account and optimized at the same time as the rest of the components of the whole plan. To the authors' knowledge, this is widely missing from the existing literature, requiring more serious attention to it in the relevant future studies.

\bibliographystyle{IEEEtran}
\bibliography{References}
\section*{Acronyms}
\begin{acronym}[MDBA-Pareto]
    \acro{1D}{One-Dimensional}
    \acro{2D}{Two-Dimensional}
    \acro{3D}{Three-Dimensional}
    \acro{6D}{Six-Dimensional}
    \acro{AAOG}{Asassembly \ac[4]{AOG}}
    \acro{AC}{Ant Colony}
    \acro{ACO}{\ac[4]{AC} Optimization}
    \acro{BC}{Bee Colony}
    \acro{BCO}{\ac[4]{BC} Optimization}
    \acro{A/DML}{Assembly/Disassembly \ac[4]{ML}}
    \acro{AI}{Artificial Intelligence}
    \acro{AIS}{Advanced Immune System}
    \acro{AL}{Assembly Line}
    \acro{ALBP}{\ac[4]{AL} \ac[4]{BP}}
    \acro{AOG}{And/Or Graph}
    \acro{AR}{Augmented Reality}
    \acro{AS}{Assembly Sequence}
    \acro{ASAGA}{Adaptive \ac[4]{SA} \ac[4]{GA}}
    \acro{ASG}{\ac[4]{AS} Generation}
    \acro{A/D-SP}{Assembly/Disassembly Sequence Planning}
    \acro{ASP}{\ac[4]{AS} Planning}
    \acro{ASTP}{Assembly \ac[4]{STP}}
    \acro{AtO}{Assemble to Order}
    \acro{BACO}{Bidirectional \ac[4]{ACO}}
    \acro{BCC}{Body Centered Cubic}
    \acro{BCI}{Brain-Computer Interface}
    \acro{BFS}{Best-First Search}
    \acro{BIM}{Building Information Modelling}
    \acro{BP}{Balancing Problem}
    \acro{CA}{Computer-Aided}
    \acro{CAAD}{\ac[4]{CA} Architectural Design}
    \acro{CAD}{\ac[4]{CA} Design}
    \acro{CAM}{\ac[4]{CA} Manufacturing}
    \acro{CAPP}{\ac[4]{CA} \ac[4]{PP}}
    \acro{CARA}{Click and Assemble \ac[4]{RA}}
    \acro{CAS}{Complex Autonomous System}
    \acro{CASTP}{Cooperative \ac[4]{ASTP}}
    \acro{CATIA}{\ac[4]{CA} Three-dimensional Interactive Application}
    \acro{CCU}{Central Control Unit}
    \acro{CDSTP}{Cooperative \ac[4]{DSTP}}
    \acro{CE}{Circular Economy}
    \acro{CIM}{Computer-Integrated Manufacturing}
    \acro{CNC}{Computer Numerical Control}
    \acro{CNN}{Convolutional Neural Network}
    \acro{CPH}{Components Per Hour}
    \acro{CPPS}{Cyber-Physical Production System}
    \acro{CPR}{Cable-driven Parallel Robot}
    \acro{CPS}{Cyber-Physical System}
    \acro{CPU}{Central Processing Unit}
    \acro{CSP}{Constraint Satisfaction Problem}
    \acro{DAOG}{Disassembly \ac[4]{AOG}}
    \acro{DAWM}{Discretely-Assembled Walking Motor}
    \acro{DCB}{Disconnecting Circuit Breaker}
    \acro{DCI}{Disassembly Cost Index}
    \acro{DDI}{Disassembly Demand Index}
    \acro{DfA}{Design for Assembly}
    \acro{DfD}{Design for Disassembly}
    \acro{DfEoL}{Design for \ac[4]{EoL}}
    \acro{DfR}{Design for Remanufacturing}
    \acro{DHGM}{Disassembly Hybrid Graph Model}
    \acro{DHI}{Disassembly Handling Index}
    \acro{DIG}{Disassembly Interference Graph}
    \acro{DL}{Disassembly Line}
	\acro{DLBP}{\ac[4]{DL} \ac[4]{BP}}
	\acro{DoF}{Degrees of Freedom}
	\acro{DOG}{Disassembly Order Graph}
	\acro{DOI}{Disassembly Operation Index}
	\acro{DP}{Dynamic Programming}
	\acro{DPD}{Delay Product Differentiation}
	\acro{DRA}{Discrete \ac[4]{RA}}
	\acro{DRL}{Deep \ac[4]{RL}}
	\acro{DS}{Disassembly Sequence}
	\acro{DSTP}{Disassembly \ac[4]{STP}}
	\acro{DSP}{\ac[4]{DS} Planning}
	\acro{DSS}{Decision Support System}
	\acro{DTW}{Dynamic Time Warping}
	\acro{EA/DML}{Educational \ac[4]{A/DML}}
	\acro{EE}{End-Effector}
	\acro{EEG}{ElectroEncephaloGraphy}
	\acro{EF-M}{Enhanced Function-Means}
	\acro{EoL}{End-of-Life}
	\acro{EV}{Exploded View}
	\acro{EVD}{\ac[4]{EV} Diagram}
	\acro{EVG}{\ac[4]{EV} Generation}
	\acro{FEA}{Finite Element Analysis}
	\acro{FPCA}{Functional Principal Component Analysis}
	\acro{GA}{Genetic Algorithm}
	\acro{GEA}{Games and Entertainment Artifacts}
	\acro{GWO}{Grey Wolf Optimization}
	\acro{HAWA}{Hybrid Ant–Wolf Algorithm}
	\acro{HCBA}{Hybrid Cuckoo–Bat Algorithm}
	\acro{HitL}{Human-in-the-Loop}
	\acro{HPN}{Hybrid \ac[4]{PN}}
	\acro{HRC}{Human-Robot Collaboration}
	\acro{HRCA}{\ac[4]{HRC} for Assembly}
	\acro{HRCD}{\ac[4]{HRC} for Disassembly}
	\acro{IC}{Integrated Circuit}
	\acro{IDP}{Inverse \ac[4]{DP}}
	\acro{IIoT}{Industrial \ac[4]{IoT}}
	\acro{IKP}{Inverse Kinematic Problem}
	\acro{IoE}{Internet of Everything}
	\acro{IoT}{Internet of Things}
	\acro{IPS}{Industrial Path Solutions}
	\acro{KT}{Knowledge Transfer}
	\acro{LbD}{Learning by Demonstration}
	\acro{LC}{Life Cycle}
	\acro{LCA}{\ac[4]{LC} Assessment}
	\acro{LLL}{Left–Left–Left}
	\acro{LLR}{Left–Left–Right}
	\acro{LoD}{Level Of Development}
	\acro{LRTDP}{Labeled Real-Time \ac[4]{DP}}
	\acro{L-system}{Lindenmayer system}
	\acro{M2M}{Machine to Machine}
	\acro{MBGA}{Many-objective Best-order-sort \ac[4]{GA}}
	\acro{MDBA-Pareto}{Modified Discrete Bees Algorithm based on Pareto}
	\acro{MDoF}{Multi \ac[4]{DoF}}
	\acro{MDPs}{Markov Decision Processes}
	\acro{MILP}{Mixed Integer Linear Programming}
	\acro{MJPN}{Median-Joining Phylogenetic Network}
	\acro{ML}{Mechatronics Line}
	\acro{MO}{Multi-Objective}
	\acro{MOO}{\ac[4]{MO} Optimization}
	\acro{MOOP}{\ac[4]{MOO} Problem}
	\acro{MR}{Multi-Robot}
	\acro{MRAL}{\ac[4]{MR} \ac[4]{AL}}
	\acro{MRDL}{\ac[4]{MR} \ac[4]{DL}}
	\acro{MRR}{Modular Reconfigurable Robot}
	\acro{MTM}{Methods \ac[4]{TM}}
	\acro{NPV}{Net Present Value}
	\acro{NLP}{Natural Language Processing}
	\acro{NP}{Non-deterministic Polynomial-time}
	\acro{NS}{Non-dominated Sorting}
	\acro{OASP}{Optimal \ac[4]{ASP}}
	\acro{ODSP}{Optimal \ac[4]{DSP}}
	\acro{OEM}{Original Equipment Manufacturer}
	\acro{OSA}{One-Side Assembly}
	\acro{PaP}{Pick-and-Place}
	\acro{PCB}{Printed Circuit Board}
	\acro{PCI}{Platforms Customisation Index}
	\acro{PD}{Parametric Design}
	\acro{PE}{Profit-oriented and Energy-efficient}
	\acro{PEASP}{\ac[4]{PE} \ac[4]{ASP}}
	\acro{PEDSP}{\ac[4]{PE} \ac[4]{DSP}}
	\acro{PN}{Petri Net}
	\acro{pose}{position and orientation}
	\acro{PP}{Process Planning}
	\acro{PRI}{Platforms Reconfiguration Index}
	\acro{QC}{Quality Control}
	\acro{RA}{Robotic Assembly}
	\acro{RAD}{Robotic Assembly and Disassembly}
	\acro{RAM}{Random-Access Memory}
	\acro{RC}{Reinforced Concrete}
	\acro{RD}{Robotic Disassembly}
	\acro{RFID}{Radio-Frequency IDentification}
	\acro{RL}{Reinforcement Learning}
	\acro{RM}{Robotic Manipulator}
	\acro{RP}{Rapid Prototyping}
	\acro{RRT}{Rapidly-exploring Random Trees}
	\acro{SA}{Simulated Annealing}
	\acro{SAAS}{Set of All \acp[4]{AS}}
	\acro{SADS}{Set of All \acp[4]{DS}}
	\acro{SAML}{Smart Assembly \ac[4]{ML}}
	\acro{SC}{Simple Cubic}
	\acro{SCARA}{Selective Compliance Assembly Robot Arm}
	\acro{SDK}{Software Development Kit}
	\acro{SDR}{Set of Directions of Removal}
	\acro{SEMS}{Smart ElectroMechanical System}
	\acro{SHPN}{Synchronized Hybrid \ac[4]{PN}}
	\acro{SM}{Smart Manufacturing}
	\acro{SMD}{Surface Mount Device}
	\acro{SMT}{Surface Mount Technology}
	\acro{SO}{Single-Objective}
	\acro{SP}{Sequence Planning}
	\acro{SOO}{\ac[4]{SO} Optimization}
	\acro{SOOP}{\ac[4]{SOO} Problem}
	\acro{SR}{Single-Robot}
	\acro{SS}{Scatter Search}
	\acro{STP}{Sequence and Task Planning}
	\acro{SWOT}{Strengths, Weaknesses, Opportunities, and Threats}
	\acro{TaMP}{Task and Motion Planning}
	\acro{TAOG}{Transformed \ac[4]{AOG}}
	\acro{Tensegrity}{Tensional integrity}
	\acro{TM}{Time Measurement}
	\acro{TMU}{\ac[4]{TM} Units}
	\acro{TNM}{TheNewMakers}
	\acro{TPD}{Task Precedence Diagram}
	\acro{TP-GMM}{Task Parametrized Gaussian Mixture Model}
	\acro{TPN}{Timed \ac[4]{PN}}
	\acro{VA}{Virtual Assembly}
	\acro{VD}{Virtual Disassembly}
	\acro{VR}{Virtual Reality}
	\acro{WD}{Wavelet Decomposition}
	\acro{WEEE}{Waste Electrical and Electronic Equipment}
	\acro{WMR}{Wheeled Mobile Robot}
	\acro{WRS}{World Robot Summit}
	\acro{WS}{WorkStation}
\end{acronym}
\end{document}